\documentclass[10pt,journal,compsoc]{IEEEtran}

\ifCLASSOPTIONcompsoc
  \usepackage[nocompress]{cite}
\else
  \usepackage{cite}
\fi

\usepackage{epsfig}
\usepackage{graphicx}
\usepackage{amsmath}
\usepackage{amssymb}
\usepackage{caption}
\usepackage{bm}
\usepackage{subfig}
\usepackage{booktabs}
\usepackage{color}
\usepackage{caption} % for color of caption
\usepackage{xcolor} % for color of caption
\usepackage{booktabs} % for color of table
\usepackage{colortbl} % for color of table
\usepackage{verbatim} % for comments
\usepackage{url}
\usepackage{algorithm}
\usepackage{algpseudocode}

\usepackage{setspace} % for the high of the table
\usepackage{ragged2e} % for alignment of abstract
\usepackage{pdfpages}

\usepackage{array} 
\makeatletter
\newcommand{\myhline}{%
    \noalign {\ifnum 0=`}\fi \hrule height 1.3pt
    \futurelet \reserved@a \@xhline
}

\newcommand{\myhlineII}{%
    \noalign {\ifnum 0=`}\fi \hrule height 1pt
    \futurelet \reserved@a \@xhline
}

\newcommand{\myhlineIII}{%
    \noalign {\ifnum 0=`}\fi \hrule height 0.95pt
    \futurelet \reserved@a \@xhline
}

\newcolumntype{"}{@{\hskip\tabcolsep\vrule width 1pt\hskip\tabcolsep}}
\makeatother

\begin{document}

\title{DVG-Face: Dual Variational Generation for Heterogeneous Face Recognition}

\author{Chaoyou~Fu,
        Xiang~Wu,
        Yibo~Hu,
        Huaibo~Huang,
        and~Ran~He,~\IEEEmembership{Senior Member,~IEEE}% <-this % stops a space
\IEEEcompsocitemizethanks{\IEEEcompsocthanksitem 
C. Fu, X. Wu, Y. Hu, H. Huang, and R. He are with the National Laboratory of Pattern Recognition, CASIA, Center for Research on Intelligent Perception and Computing, CASIA, Center for Excellence in Brain Science and Intelligence Technology, CAS, and School of Artificial Intelligence, University of Chinese Academy of Sciences, Beijing 100190, China.
E-mail: \{chaoyou.fu, rhe\}@nlpr.ia.ac.cn, \{alfredxiangwu, huyibo871079699\}@gmail.com, huaibo.huang@cripac.ia.ac.cn.
(Corresponding author: Ran He.)
}% <-this % stops an unwanted space
}

\markboth{Journal of \LaTeX\ Class Files}%
{Shell \MakeLowercase{\textit{et al.}}: Bare Demo of IEEEtran.cls for Computer Society Journals}

\IEEEtitleabstractindextext{%
\begin{abstract}
\justifying
Heterogeneous Face Recognition (HFR) refers to matching cross-domain faces and plays a crucial role in public security. Nevertheless, HFR is confronted with challenges from large domain discrepancy and insufficient heterogeneous data. In this paper, we formulate HFR as a dual generation problem, and tackle it via a novel Dual Variational Generation (DVG-Face) framework. Specifically, a dual variational generator is elaborately designed to learn the joint distribution of paired heterogeneous images. However, the small-scale paired heterogeneous training data may limit the identity diversity of sampling. In order to break through the limitation, we propose to integrate abundant identity information of large-scale visible data into the joint distribution. Furthermore, a pairwise identity preserving loss is imposed on the generated paired heterogeneous images to ensure their identity consistency. As a consequence, massive new diverse paired heterogeneous images with the same identity can be generated from noises. The identity consistency and identity diversity properties allow us to employ these generated images to train the HFR network via a contrastive learning mechanism, yielding both domain-invariant and discriminative embedding features. Concretely, the generated paired heterogeneous images are regarded as positive pairs, and the images obtained from different samplings are considered as negative pairs. Our method achieves superior performances over state-of-the-art methods on seven challenging databases belonging to five HFR tasks, including NIR-VIS, Sketch-Photo, Profile-Frontal Photo, Thermal-VIS, and ID-Camera. The related code will be released at~{\color{blue} \url{https://github.com/BradyFU}}.
\end{abstract}

}

\maketitle

\IEEEdisplaynontitleabstractindextext

\IEEEpeerreviewmaketitle

\IEEEraisesectionheading{\section{Introduction}\label{introduction}}

\IEEEPARstart{I}{n} the last two decades, face recognition for visible (VIS) images has made a great breakthrough using deep convolutional neural networks \cite{wen2016discriminative,deng2019arcface}. 
Nevertheless, in some practical security applications, a face recognition network often needs to match cross-domain face images rather than merely VIS, raising the problem of Heterogeneous Face Recognition (HFR).
For example, the near-infrared (NIR) imaging sensor has been integrated into many mobile devices, because it offers an inexpensive and effective solution to obtain clear NIR images in extreme lightings or even dark environment.
At the same time, the enrollments of face templates are usually VIS images, which requires the recognition network to match the heterogeneous NIR-VIS images.
Unfortunately, due to the large domain discrepancy, the performance of the recognition network trained on VIS images often degrades dramatically in such a heterogeneous case \cite{DBLP:journals/corr/abs-1708-02412}. Other HFR tasks include Sketch-Photo \cite{bhatt2012memetic}, Profile-Frontal Photo \cite{gross2010multi}, Thermal-VIS \cite{panetta2018comprehensive}, and ID-Camera \cite{huo2017heterogeneous}. 
In order to bridge the domain discrepancy between the heterogeneous data, extensive research efforts have been undertaken to match cross-domain features \cite{ouyang2016survey}.
However, due to the difficulty of data acquisition, it is usually not feasible to collect large-scale heterogeneous databases.
Over-fitting often occurs when the recognition network is trained on such insufficient data \cite{DBLP:journals/corr/abs-1708-02412}.

\begin{figure*}[t]
\centering
\includegraphics[width=0.985\textwidth]{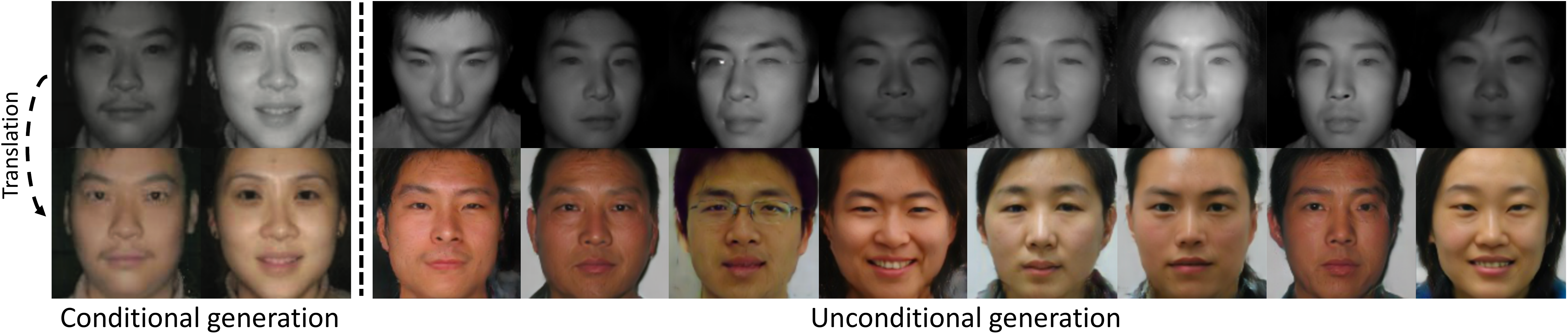}
\caption{Comparisons between conditional image-to-image generation \cite{Song2018AdversarialDH} (on the left, the top is the input NIR and the bottom is the generated VIS) and our proposed unconditional dual variational generation (on the right, all paired NIR-VIS images are generated from noises). For the conditional generation, given one NIR image, the generator can only synthesize one VIS image with the same attributes (\textit{e.g.} the pose and the expression) except for the spectrum. By contrast, DVG-Face has the ability to generate large-scale new paired heterogeneous images with abundant intra-class diversity from noises.
}
\label{fig_2}
\end{figure*}

With the rapid development of deep generative models, ``recognition via generation'' has been a hot research topic in the computer vision community \cite{Huang2017BeyondFR}.
Deep generative models are good at learning the mapping among different domains, and are therefore usually used to reduce the domain discrepancy \cite{isola2017image}.
For example, \cite{Song2018AdversarialDH} proposes to transfer NIR images to VIS ones via a two-path generative model. \cite{Zhang2018IJCV} introduces a multi-stream feature fusion manner with Generative Adversarial Networks (GANs) \cite{Goodfellow2014GenerativeAN} to synthesize photo-realistic VIS images from polarimetric thermal ones. 
However, these conditional image-to-image translation based methods face two challenging issues:
(1) \textbf{Diversity}. Since these methods adopt a ``one-to-one'' translation manner, the generator can only synthesize one new image of the target domain for each input image of the source domain. Hence, these methods can merely generate a limited number of new data to reduce the domain discrepancy.
Meanwhile, as exemplified in the left part of Fig.~\ref{fig_2}, the generated images have the same attributes (\textit{e.g.} the pose and the expression) as the inputs except for the spectrum, which limits the intra-class diversity of the generated images.
(2) \textbf{Consistency}. 
These methods require that the generated images belong to specific classes consistent with the inputs.
Nevertheless, it is hard to meet this requirement, because the widely adopted identity preserving loss \cite{Huang2017BeyondFR,duan2019pose} only constrains the feature distance between the generated images and the targets, while ignores both intra-class and inter-class distances.
The above two issues make it rather difficult for deep generative models to effectively boost the performance of HFR.

In order to tackle the above challenges, we propose an unconditional Dual Variational Generation (DVG-Face) framework that can generate large-scale new diverse paired heterogeneous images.
Different from the aforementioned conditional generation in HFR, unconditional generation has the ability to synthesize new data by sampling from noises \cite{Kingma2013AutoEncodingVB,karras2017progressive}.
However, the small number of the heterogeneous training data may limit the identity diversity of sampling.
In order to break through the limitation, we introduce abundant identity information of large-scale VIS images into the generated images.
By contrast, these VIS images are merely leveraged to pre-train the HFR network in previous works \cite{deng2019residual,XWu:2018}, as discussed in Section~\ref{VIS data}.
Moreover, DVG-Face focuses on the identity consistency of the generated paired heterogeneous images, rather than whom the generated images belong to. This avoids the consistency problem of the previous conditional generation based methods.
Finally, a contrastive learning mechanism is employed to take advantage of these generated unlabeled images to train the HFR network.
Specifically, benefiting from the identity consistency property, the generated paired heterogeneous images are regarded as positive pairs. Due to the identity diversity property, the images obtained from different samplings are considered as negative pairs, according to the instance discrimination criterion \cite{wu2018unsupervised,bachman2019learning}.

The framework of our DVG-Face is depicted in Fig.~\ref{fig_framework}.
We first train the dual variational generator with both paired heterogeneous data and large-scale unpaired VIS data.
In the latent space, the joint distribution of paired heterogeneous data is disentangled into identity representations and domain-specific attribute distributions, which is implemented by an angular orthogonal loss.
Fig.~\ref{fig_3} presents the disentanglement results.
This disentanglement enables the integration of abundant identity information of the unpaired VIS images into the joint distribution.
In the pixel space, we impose a pairwise identity preserving loss on the generated paired heterogeneous images to guarantee their identity consistency.
After the training of the dual variational generator, we can obtain large-scale new diverse paired heterogeneous images with the same identity by sampling from noises.
These generated images are organized as positive and negative pairs, and then are fed into the HFR network.
A contrastive learning mechanism further facilitates the HFR network to learn both domain-invariant and discriminative embedding features.

The main contributions are summarized as follows:
\begin{itemize}
  \item We provide a new insight into HFR, considering it as a dual generation problem. This results in a novel unconditional Dual Variational Generation (DVG-Face) framework, which samples large-scale new diverse paired heterogeneous data from noises to boost the performance of HFR.

  \item Abundant identity information is integrated into the joint distribution to enrich the identity diversity of the generated data. Meanwhile, a pairwise identity preserving loss is imposed on the generated paired images to ensure their identity consistency. These two properties allow us to make better use of the generated unlabeled data to train the HFR network.
  
  \item By regarding the generated paired images as positive pairs and the images obtained from different samplings as negative pairs, the HFR network is optimized via contrastive learning to learn both domain-invariant and discriminative embedding features.

  \item Extensive experiments on seven HFR databases demonstrate that our method significantly outperforms state-of-the-art methods. In particular, on the challenging low-shot Oulu-CASIA NIR-VIS database, we improve the best TPR@FPR=10$^{-5}$ by 29.2$\%$. Besides, compared with the baseline trained on the Tufts Face database, VR@FAR=$1\%$ is increased by 26.1$\%$ after adding the generated data.
\end{itemize}

This paper is an extension of our previous conference version \cite{fu2019dual}, and there are three major improvements over the preliminary one:
(1) \textbf{The generated images have richer identity diversity}.
For the preliminary version, the generator can only be trained with the small-scale paired heterogeneous data, thereby limiting the identity diversity of the generated images.
In the current version, the architecture and the training manner of the generator are redesigned, allowing it to be trained with both paired heterogeneous data and large-scale unpaired VIS data.
The introduction of the latter greatly enriches the identity diversity of the generated images.
(2) \textbf{The generated images are leveraged more efficiently}.
The preliminary version trains the HFR network with the generated paired data via a pairwise distance loss, resorting to the identity consistency property.
On this basis, benefiting from the aforementioned identity diversity property, the current version further regards the images obtained from different samplings as negative pairs, formulating a contrastive learning mechanism.
Hence, the preliminary version can only leverage the generated images to reduce the domain discrepancy, while the current version utilizes the generated images to learn both domain-invariant and discriminative embedding features.
(3) \textbf{More insightful analyses and more experiments are added}. Apart from NIR-VIS and Sketch-Photo, we further explore Profile-Frontal Photo, Thermal-VIS, and ID-Camera HFR tasks.
Moreover, the current version gains significant improvements over the preliminary one on all the databases.

\section{Related Work}
\subsection{Heterogeneous Face Recognition}
HFR has attracted increasing attention of researchers for its crucial practical value.
In this subsection, we review the development of HFR from the perspectives of feature-level learning and image-level learning.

\emph{Feature-level learning} based methods aim at seeking discriminative feature representations.
Some methods try to capture a common latent space between heterogeneous data for relevance measurement.
\cite{klare2012heterogeneous} employs a prototype random subspace to improve the HFR performance.
\cite{shao2014generalized} introduces a generalized transfer subspace with a low-rank constraint.
There are also some methods that focus on learning domain-invariant features.
Traditional works mainly use handcrafted features. For example, \cite{Klare2011MatchingFS} extracts Histograms of Oriented Gradients (HOG) features with sparse representations to obtain cross-domain semantics.
\cite{gong2017heterogeneous} proposes a common encoding model to obtain discriminant information for HFR.
Recently, the thriving deep learning has powerful feature extraction capability, and thus has been widely applied in HFR.
\cite{saxena2016heterogeneous} explores multiple different metric learning methods to reduce the domain gap.
\cite{deng2019residual} designs a residual compensation framework that learns different modalities in separate branches.
However, due to the lack of large-scale heterogeneous databases, the deep learning based methods often tend to over-fit \cite{XWu:2019}. In order to alleviate the over-fitting problem, our method generates large-scale new diverse paired data as an augmentation, thus significantly boosting the recognition performance.

\emph{Image-level learning} based methods mainly operate on the pixel space to improve the performance of HFR.
\cite{Tang2003FaceSS,peng2015multiple,peng2015superpixel} focus on Sketch-Photo synthesis to reduce the domain gap.
\cite{huang2013coupled} improves the recognition performance by using coupled dictionary learning for cross-domain image synthesis.
\cite{hu2017frankenstein} proposes to composite facial parts, \textit{e.g.} eyes, to augment the small database to a larger one.
\cite{Lezama2017NotAO} uses a CNN to perform a cross-spectral hallucination, reducing the domain gap in pixel space.
Recently, the rapidly developed GANs \cite{Goodfellow2014GenerativeAN} provide a solution to the problem of HFR.
\cite{cao2018data} utilizes image-to-image translation methods to synthesize new data, and then incorporates the synthesized data into the training set to augment the intra-class scale and reduce inter-class diversity.
\cite{zhang2019cascaded} proposes a cascaded face sketch synthesis model to deal with the illumination problem.
As discussed in Section 1, the image-to-image translation based methods are limited in diversity and consistency. In order to break through the two limitations, DVG-Face employs a novel unconditional generation manner to generate large-scale new diverse paired images from noises, and a new pairwise identity preserving loss to ensure the identity consistency of the generated paired heterogeneous images.

\begin{figure*}[t]
\centering
\includegraphics[width=0.985\textwidth]{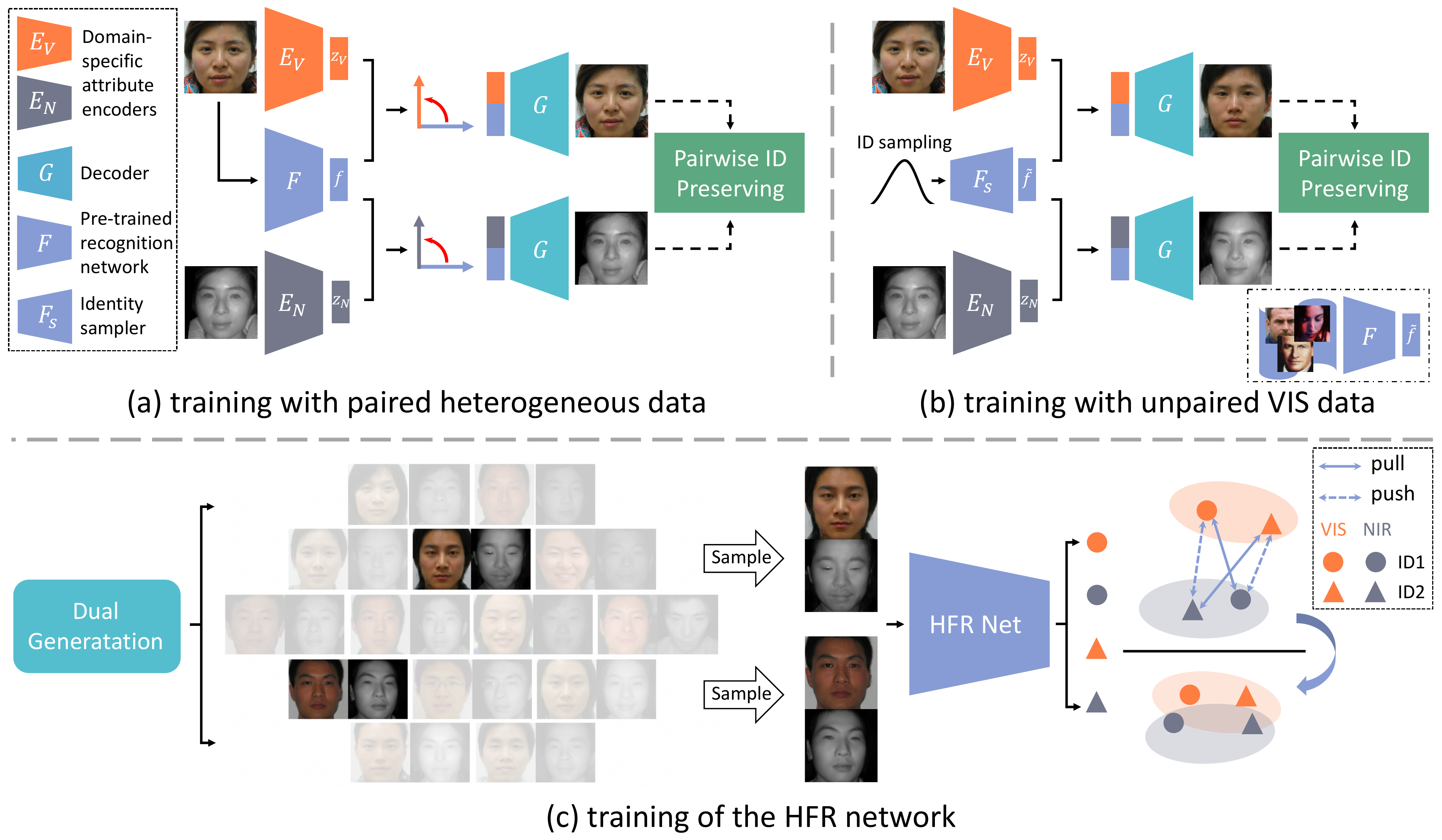}
\caption{Illustration of our DVG-Face. 
(a) and (b): The training of the dual variational generator involves both paired heterogeneous data and unpaired VIS data (MS-Celeb-1M~\cite{guo2016ms}). 
The former is used to disentangle identity and attribute representations.
The latter is introduced to enrich the identity diversity of the generated images.
The identity representations of the latter can be obtained from either the identity sampler $F_s$ or the pre-trained face recognition network $F$, as discussed in Section~\ref{identity sampling}.
In addition, a pairwise identity preserving loss is imposed on the generated paired images to guarantee their identity consistency.
(c): After training the generator, we leverage it to synthesize a great deal of new paired heterogeneous images.
Benefiting from the identity consistency property, the generated paired images are regarded as positive pairs.
Due to the identity diversity property, the images obtained from different samplings are considered as negative pairs.
A contrastive learning mechanism is imposed on the HFR network, yielding both domain-invariant and discriminative embedding features.
}
\label{fig_framework}
\end{figure*}

\subsection{Generative Model}
As a hot topic in machine learning and computer vision communities, generative model has made gratifying progress in recent years\cite{Goodfellow2014GenerativeAN,Kingma2013AutoEncodingVB,van2016conditional,kingma2018glow}.
Usually, it can be divided into two types: unconditional generative model and conditional generative model.

\emph{Unconditional generative model} synthesizes data from noises.
The most prominent models include Variational AutoEncoders (VAEs) \cite{Kingma2013AutoEncodingVB} and Generative Adversarial Networks (GANs) \cite{Goodfellow2014GenerativeAN}.
VAEs adopt variational inference to lean data distribution, and consist of an encoder network $q_{\phi}(z|x)$ and a decoder network $p_{\theta}(x|z)$.
The Evidence Lower Bound Objective (ELBO) is derived for an approximate optimization:
\begin{equation}
\label{eq:vae}
\resizebox{0.49\textwidth}{!}{
    $\log p_{\theta}(x) \geq \mathbb{E}_{q_{\phi}(z|x)}\log p_{\theta}(x|z) - D_{\text{KL}}(q_{\phi}(z|x) || p(z))$,
    }
\end{equation}
where the first term is a reconstruction loss and the second term is a Kullback-Leibler divergence.
$p(z)$ is usually set to a standard Gaussian distribution.
Recently, some variants are proposed to improve synthesis quality.
IntroVAE \cite{Huang2018IntroVAEIV} generates high-resolution images via an introspective manner.
VQ-VAE \cite{van2017neural} obtains high-quality images via learning discrete representations with an autoregressive prior. 

In a different way, GANs leverage an adversarial mechanism to learn data distribution implicitly.
Generally, GANs consist of two networks: a generator $G$ and a discriminator $D$.
$G$ generates data from a prior $p(z)$ to confuse $D$, while $D$ is trained to distinguish the generated data from the real data.
Such an adversarial mechanism is formulated as:
\begin{equation}
\label{eq:gan}
\resizebox{0.49\textwidth}{!}{
    $\min \limits_{G} \max \limits_{D} \mathbb{E}_{x \sim p_{data}(x)} \left [ \log D(x)  \right ] + \mathbb{E}_{z \sim p_{z}(z)} \left [ \log(1 - D(G(z))) \right ]$.
    }
\end{equation}
GANs are good at generating high-quality images.
PGGAN \cite{karras2017progressive} significantly improves the resolution of the synthesized images to 1024$\times$1024.
SinGAN \cite{shaham2019singan} generates realistic images merely using one single training image.
CoGAN \cite{liu2016coupled} concerns a problem similar to the one tackled by our method. A weight-sharing manner is adopted to generate paired data from two domains. However, due to the lack of explicit identity constrains, it is rather challenging for such a weight-sharing manner of CoGAN to generate paired data with the same identity, as discussed in Section~\ref{identity consistency}.

\emph{Conditional generative model} synthesizes data according to the given conditions.
pix2pix \cite{isola2017image} realizes photo-realistic paired image-to-image translation with a conditional generative adversarial loss.
CycleGAN \cite{zhu2017unpaired} employs a cycle-consistent network to tackle the unpaired image translation problem.
GAN Compression \cite{li2020gan} reduces the model size and the inference time of Conditional GANs by the ways of Knowledge Distillation (KD) and Neural Architecture Search (NAS).
MUNIT \cite{huang2018multimodal} decomposes image representations into a content code and a style code, realizing exemplar based image synthesis.
BigGAN \cite{brock2018large} is the pioneer work for high-resolution natural image synthesis.
StarGAN \cite{choi2019stargan} is designed for unpaired multi-domain image-to-image translation.
StyleGAN \cite{karras2019style} proposes a style-based generator that can separate facial attributes automatically.
SPADE \cite{park2019semantic} synthesizes high-quality images with the guidance of semantic layouts.
InterFaceGAN \cite{shen2019interpreting} edits face images according to the interpretable latent semantics learned by GANs.
In-Domain GAN inversion \cite{zhu2020domain} proposes to invert a pre-trained GAN model for real image editing, resorting to an image reconstruction in the pixel space and a semantic regularization in the latent space.

\begin{table}[t]
    \centering
    \caption{Meaning of the symbols used in our method.}
    \label{table-symbol}
    \begin{spacing}{0.9}
    \resizebox{0.45\textwidth}{!}{
    \begin{tabular}{c|l}
        Symbol & Meaning \\
        \myhlineII
        $E_N$/$E_V$ & Attribute encoder for NIR/VIS \\
        $G$ & Decoder network \\
        $F$ & Pre-trained face recognition network \\
        $F_s$ & Pre-trained identity sampler \\
        \hline
        $I_N$/$I_V$ & NIR/VIS image \\
        $z_N$/$z_V$ & Attribute distribution of $I_N$/$I_V$ \\
        $f$ & Identity representation of $I_V$ \\
        $\hat{I}_N$/$\hat{I}_V$ & Reconstructed NIR/VIS image \\
        $\hat{f}_N$/$\hat{f}_V$ & Identity representation of $\hat{I}_N$/$\hat{I}_V$ \\
        $\tilde{f}$ & Identity representation sampled from $F_s$ \\
        $\tilde{I}_N$/$\tilde{I}_V$ & Generated NIR/VIS image with $\tilde{f}$ \\
        $\tilde{f}_N$/$\tilde{f}_V$ & Identity representation of $\tilde{I}_N$/$\tilde{I}_V$ \\
    \end{tabular}}
    \end{spacing}
\end{table}

\section{Method} \label{method}
The goal of our method is to generate a large number of paired heterogeneous data from noises to boost the performance of HFR.
Correspondingly, our method is divided into two parts to tackle the problems of (1) \emph{how to generate diverse paired heterogeneous data} and (2) \emph{how to effectively take advantage of these generated data}. 
These two parts are introduced in the following two subsections, respectively.
We take NIR-VIS as an example for illustration. Other heterogeneous face images are in the same way.

\subsection{Dual Generation} \label{dual generation}
As stated in Section~\ref{introduction}, we expect the generated paired images to have two properties: identity diversity and identity consistency.
On the one hand, in order to promote the identity diversity, we adopt an unconditional generation manner because it has the ability to generate new samples from noises \cite{Kingma2013AutoEncodingVB,karras2017progressive}. 
Meanwhile, considering that the small number of the paired heterogeneous training data may limit the identity diversity of the sampling, we introduce abundant identity information of large-scale VIS data into the generated images.
On the other hand, in order to constrain the identity consistency property, we impose a pairwise identity preserving loss on the generated paired images, minimizing their feature distances in the embedding space.

The above ideas are implemented by a dual variational generator and its framework is shown in Figs.~\ref{fig_framework} (a) and (b).
The generator contains two domain-specific attribute encoders $E_N$ and $E_V$, a decoder $G$, a pre-trained face recognition network $F$, and an identity sampler $F_s$.
Among them, $E_N$ and $E_V$ are utilized to learn domain-specific attribute distributions of NIR and VIS data, respectively.
$F$ is used to extract identity representations.
$F_s$ can flexibly sample abundant identity representations from noises.
The joint distribution of the paired heterogeneous data consists of the identity representations and the attribute distributions.
$G$ maps the joint distribution to the pixel space.
The parameters of $F$ and $F_s$ are fixed, while those of the other networks are updated during the training with both paired heterogeneous data and unpaired VIS data.
Table~\ref{table-symbol} summarizes the meaning of the symbols used in our method.

\subsubsection{Training with Paired Heterogeneous Data}
Given a pair of NIR-VIS data $I_N$ and $I_V$ with the same identity, the dual variational generator learns disentangled joint distribution in the latent space.
Specifically, a face recognition network pre-trained on MS-Celeb-1M \cite{guo2016ms} is adopted as the feature extractor $F$.
The compact embedding features extracted by $F$ are thought to be only identity related \cite{bao2018towards}.
Meanwhile, considering that $F$ is better at extracting the features of VIS images than those of NIR images, we use $f = F(I_V)$ as the identity representation of both $I_N$ and $I_V$.
Then, the encoders $E_N$ and $E_V$ are leveraged to learn domain-specific attribute distributions $z_N = q_{\phi_N}(z_N|I_N)$ and $z_V = q_{\phi_V}(z_V|I_V)$, respectively. 
$\phi_N$ and $\phi_V$ are the parameters of the encoders.
According to the reparameterization trick \cite{Kingma2013AutoEncodingVB}, $z_N = u_N + \sigma_N \odot \epsilon$ and $z_V = u_V + \sigma_V \odot \epsilon$, where $u$ denotes mean and $\sigma$ denotes standard deviation.
$\odot$ denotes Hadamard product and $\epsilon$ is a noise sampled from a multi-variate standard Gaussian distribution.
Subsequently, in order to ensure that $z_N$ and $z_V$ are merely attribute related, we impose an angular orthogonal loss between the attribute and the identity representations.
Finally, the disentangled identity and attribute representations constitute the joint distribution of paired NIR-VIS data, and then are fed to the decoder $G$ to reconstruct the inputs $I_N$ and $I_V$.

A total of four loss functions are involved in the above process, including an angular orthogonal loss, a distribution learning loss, a pairwise identity preserving loss, and an adversarial loss. These losses are introduced below.

\textbf{Angular Orthogonal Loss.}
The angular orthogonal loss is imposed between $z_N$ and $f$ and between $z_V$ and $f$.
For the $L_2$ normalized $z_V$ and $f$, their cosine similarity is formulated as:
\begin{equation}\label{cos}
    \text{cos}(z_V, f) = \langle z_V, f \rangle,
\end{equation}
where $\langle \cdot, \cdot \rangle$ denotes inner product.
When the two representations are orthogonal, Eq.~(\ref{cos}) is equal to zero. Minimizing the absolute value of Eq.~(\ref{cos}) will force $z_V$ and $f$ to be orthogonal, and thus they are disentangled.
Ultimately, the angular orthogonal loss that considers both normalized $z_N$ and $z_V$ is defined as:
\begin{equation} \label{orthogonal}
    \mathcal{L}_{\text{ort}} = \left | \langle z_V, f \rangle \right | + \left | \langle z_N, f \rangle \right |.
\end{equation}
In the following parts, $z_N$, $z_V$, and the output of $F(\cdot)$, \textit{e.g.} $f = F(I_V)$, all denote the normalized ones.

\textbf{Distribution Learning Loss.}
Inspired by VAEs \cite{Kingma2013AutoEncodingVB}, the posterior distributions $q_{\phi_N}(z_N|I_N)$ and $q_{\phi_V}(z_V|I_V)$ are constrained by a Kullback-Leibler divergence:
\begin{equation} \label{kl}
    \mathcal{L}_{\text{kl}} = D_{\text{KL}}(q_{\phi_N}(z_N|x_N) || p(z_N)) + D_{\text{KL}}(q_{\phi_V}(z_V|x_V) || p(z_V)),
\end{equation}
where both the priors $p(z_N)$ and $p(z_V)$ are multi-variate standard Gaussian distributions.
After obtaining the identity representation $f$ as well as the domain-specific attribute representations $z_N$ and $z_V$, the decoder network is required to reconstruct the inputs $I_N$ and $I_V$:
\begin{equation}
    \mathcal{L}_{\text{rec}} = - \mathbb{E}_{q_{\phi_N}(z_N|I_N)\cup q_{\phi_V}(z_V|I_V)} \log p_{\theta}(I_N, I_V|f, z_N, z_V).
\end{equation}
In practice, $\mathcal{L}_{\text{rec}}$ is implemented as:
\begin{equation} \label{rec}
    \mathcal{L}_{\text{rec}} = || \hat{I}_N - I_N ||_1 + || \hat{I}_V - I_V ||_1,
\end{equation}
where $\hat{I}_N = G(f, z_N)$ and $\hat{I}_V = G(f, z_V)$.

In general, according to \cite{Kingma2013AutoEncodingVB}, the distribution learning loss is a combination of $\mathcal{L}_{\text{kl}}$ and $\mathcal{L}_{\text{rec}}$:
\begin{equation} \label{dis}
    \mathcal{L}_{\text{dis}} = \mathcal{L}_{\text{kl}} + \mathcal{L}_{\text{rec}}.
\end{equation}

\textbf{Pairwise Identity Preserving Loss.}
In order to preserve the identity of the generated data, previous conditional generation based methods usually adopt an identity preserving loss \cite{Huang2017BeyondFR,hu2018pose,duan2019pose}. 
A pre-trained face recognition network is used to extract the embedding features of the generated data and those of the real target data respectively, and then the two features are forced to be as close as possible.
However, since there are neither intra-class nor inter-class constraints, it is challenging to guarantee that the generated images belong to specific classes consistent with the targets.

As discussed in Section~\ref{introduction}, different from the previous methods, we focus on the identity consistency of the generated paired images, rather than whom the generated images belong to.
Therefore, we propose a pairwise identity preserving loss that constrains the feature distance between $\hat{f}_N = F(\hat{I}_N)$ and $\hat{f}_V = F(\hat{I}_V)$:
\begin{equation} \label{pairwise}
    \mathcal{L}_{\text{ip-pair}} = 1 - \langle \hat{f}_N, \hat{f}_V \rangle.
\end{equation}
Minimizing Eq.~(\ref{pairwise}) will increase the cosine similarity between $\hat{f}_N$ and $\hat{f}_V$. Besides, in order to stabilize the training process, we also constrain the feature distance between $\hat{f}_N$ and $f$ and that between $\hat{f}_V$ and $f$:
\begin{equation} \label{ip-rec}
    \mathcal{L}_{\text{ip-rec}} = (1 - \langle \hat{f}_N, f \rangle) + (1 - \langle \hat{f}_V, f \rangle).
\end{equation}
The pairwise identity preserving loss that considers both $\mathcal{L}_{\text{ip-pair}}$ and $\mathcal{L}_{\text{ip-rec}}$ is formulated as:
\begin{equation} \label{ip}
\begin{split}
    \mathcal{L}_{\text{ip}} = \mathcal{L}_{\text{ip-pair}} + \mathcal{L}_{\text{ip-rec}}.
\end{split}
\end{equation}

\textbf{Adversarial Loss.}
Same as \cite{shu2018deforming}, an extra adversarial loss $\mathcal{L}_{\text{adv}}$ \cite{Goodfellow2014GenerativeAN} is also introduced to increase the sharpness of the generated data, where the dual variational generator and a discriminator are optimized alternately. We first fix the discriminator and train the generator with the adversarial loss to confuse the discriminator:
\begin{equation} \label{adv-g}
\begin{split}
    \mathcal{L}_{\text{adv-g}} = \log(1 - D(\hat{I}_N)) + \log(1 - D(\hat{I}_V)),
\end{split}
\end{equation}
where $D$ denotes the discriminator.
Then, we fix the generator and train the discriminator to distinguish the generated data from the real data:
\begin{equation} \label{adv-d}
\begin{split}
    \mathcal{L}_{\text{adv-d}} = & - \log(D(I_N)) - \log(1 - D(\hat{I}_N)) \\
                               & - \log(D(I_V)) - \log(1 - D(\hat{I}_V)).
\end{split}
\end{equation}

\textbf{Overall Loss.}
The overall loss for the training with paired heterogeneous data is a weighted sum of the above distribution learning loss $\mathcal{L}_{\text{dis}}$, the angular orthogonal loss $\mathcal{L}_{\text{ort}}$, the pairwise identity preserving loss $\mathcal{L}_{\text{ip}}$, and the adversarial loss $\mathcal{L}_{\text{adv}}$:
\begin{equation} \label{pair}
    \mathcal{L}_{\text{pair}} = \mathcal{L}_{\text{dis}} + \lambda_1 \mathcal{L}_{\text{ort}} + \lambda_2 \mathcal{L}_{\text{ip}} + \lambda_3 \mathcal{L}_{\text{adv}},
\end{equation}
where $\lambda_1$, $\lambda_2$, and $\lambda_3$ are trade-off parameters. Among them, $\lambda_3$ is fixed to 0.1 according to \cite{shu2018deforming}.

\subsubsection{Training with Unpaired VIS Data} \label{training unpair}
The number of the paired heterogeneous training data is limited, which may affect the identity diversity of the generated images. 
Therefore, we introduce abundant identity information of large-scale VIS data into the generated images.
For the acquisition of the identity information, a straightforward approach is to use a pre-trained face recognition network to extract identities from the large-scale VIS data, as shown in the lower right corner of Fig.~\ref{fig_framework} (b).
However, in this situation, if we desire to generate large-scale new paired data at the testing phase, we must have the same number of VIS data with different identities. 
This will make our framework a conditional generative model, leading to the diversity problem as mentioned in Section~\ref{introduction}.
In order to overcome this obstacle, inspired by \cite{deng2020disentangled}, we introduce an identity sampler to replace the recognition network.
Specifically, we first adopt the recognition network to extract the embedding features of MS-Celeb-1M \cite{guo2016ms}, and then leverage these embedding features to train a VAE model \cite{Kingma2013AutoEncodingVB}.
After the training, the decoder of VAE is used as the identity sampler, which can map the points in a standard Gaussian noise to identity representations, as shown in Fig.~\ref{fig_framework} (b).
Equipped with the identity sampler, our framework becomes an unconditional generative model. The required identity representations can be flexibly sampled from noises.
A detailed discussion about the face recognition network and the identity sampler is provided in Section~\ref{identity sampling}.

Since these sampled identity representations do not have corresponding ground truth paired heterogeneous images, we propose to train the generator in an unpaired way.
To begin with, we sample an identity representation $\tilde{f}$ via the identity sampler $F_s$.
Then, $\tilde{f}$ as well as the attribute representations $z_N$ and $z_V$ in Eq.~(\ref{orthogonal}) are fed into the decoder $G$.
Finally, a pair of new heterogeneous images, \textit{i.e.} $\tilde{I}_N = G(\tilde{f}, z_N)$ and $\tilde{I}_V = G(\tilde{f}, z_V)$, that does not belong to the heterogeneous database is generated.
We constrain $\tilde{I}_N$ and $\tilde{I}_V$ from the aspects of appearance and semantic.
For appearance, we introduce a small-weight reconstruction loss \cite{bao2018towards} to force the appearance of the generated images to be consistent with that of the inputs:
\begin{equation} \label{rec-u}
    \mathcal{L}_{\text{rec}}^\text{u} = \eta \left ( || \tilde{I}_N - I_N ||_1 + || \tilde{I}_V - I_V ||_1 \right ),
\end{equation}
where $\eta$ is set to $0.1$ according to \cite{bao2018towards}. 
For semantic, same as the aforementioned pairwise identity preserving manner in Eq.~(\ref{ip}), we constrain $\tilde{f}_N = F(\tilde{I}_N)$ and $\tilde{f}_V = F(\tilde{I}_V)$ via:
\begin{equation} \label{ip-u}
\begin{split}
    \mathcal{L}_{\text{ip}}^\text{u} = (1 - \langle \tilde{f}_N, \tilde{f}_V \rangle) + (1 - \langle \tilde{f}_N, \tilde{f} \rangle) + (1 - \langle \tilde{f}_V, \tilde{f} \rangle).
\end{split}
\end{equation}
Consequently, the overall loss for the training with unpaired VIS data is a weighted sum of $\mathcal{L}_{\text{rec}}^\text{u}$, $\mathcal{L}_{\text{ip}}^\text{u}$, and an extra adversarial loss $\mathcal{L}_{\text{adv}}^\text{u}$ (replacing $\hat{I}_N$/$\hat{I}_V$ in Eq.~(\ref{adv-g}) and Eq.~(\ref{adv-d}) with $\tilde{I}_N$/$\tilde{I}_V$):
\begin{equation} \label{unpair}
    \mathcal{L}_{\text{unpair}} = \mathcal{L}_{\text{rec}}^\text{u} + \lambda_2 \mathcal{L}_{\text{ip}}^\text{u} + \lambda_3 \mathcal{L}_{\text{adv}}^\text{u},
\end{equation}
where $\lambda_2$ and $\lambda_3$ are the same as those in Eq.~(\ref{pair}).
Algorithm~\ref{alg-dual} shows the training process with both paired heterogeneous data and unpaired VIS data.

\begin{algorithm}[t] 
\caption{Training process of the generator} 
\label{alg-dual} 
\begin{algorithmic}[1] 
\Require 
Heterogeneous training set $I_i, i \in \{N, V\}$; 
A pre-trained face recognition network $F$;
A pre-trained identity sampler $F_s$;
\Ensure
The parameters of the dual variational generator, including $E_N$, $E_V$, and $G$;
\For{$t = 1,...,T$} 
\State \emph{\# Training with paired heterogeneous data.}
\State Sample a batch of paired training data $\{I_N, I_V\}$;
\State Use $F$ to extract the identity feature $f = F(I_V)$;
\State Feed $I_N$, $I_V$, and $f$ to the generator;
\State Calculate the loss in Eq.~(\ref{pair});
\State \emph{\# Training with unpaired VIS data.}
\State Sample a random identity feature $\tilde{f}$ via $F_s$;
\State Feed $I_N$, $I_V$, and $\tilde{f}$ to the generator;
\State Calculate the loss in Eq.~(\ref{unpair});
\EndFor \\ 
\Return $E_N$, $E_V$, and $G$.
\end{algorithmic} 
\end{algorithm}

\subsection{Heterogeneous Face Recognition}
After the training of the dual variational generator, we first employ it to generate large-scale paired heterogeneous images, and then leverage these images to facilitate the training of the HFR network. 
For HFR, we use a face recognition network pre-trained on MS-Celeb-1M as the backbone, and train it with both the limited number of real heterogeneous data $I_i~(i \in \{N, V\})$ and the large-scale generated heterogeneous data $\tilde{I}_i~(i \in \{N, V\})$.

For the real heterogeneous data, we utilize a softmax loss to optimize the HFR network:
\begin{equation} \label{softmax}
    \mathcal{L}_{\text{cls}} = \sum_{i \in \{N, V\}} \text{softmax}(f_i, y),
\end{equation}
where $y$ is the class label of the input $I_i$ and $f_i = F(I_i)$.
$F$ denotes the pre-trained face recognition network, same as the feature extractor in Section~\ref{dual generation}. 
The difference is that the parameters of the former are updated while those of the latter are fixed.

For the generated data, since there is no specific class label, the above softmax loss is inapplicable.
However, benefiting from the properties of identity consistency and identity diversity, we propose to take advantage of these generated data via a contrastive learning mechanism \cite{hadsell2006dimensionality}.
To be specific, as depicted in Fig.~\ref{fig_framework} (c), we first randomly sample two paired heterogeneous images ($\tilde{I}_N^1$, $\tilde{I}_V^1$) and ($\tilde{I}_N^2$, $\tilde{I}_V^2$) from the generated database.
Based on the identity consistency property, the paired heterogeneous images ($\tilde{I}_N^1$, $\tilde{I}_V^1$) and ($\tilde{I}_N^2$, $\tilde{I}_V^2$) are set as positive pairs.
At the same time, thanks to the identity diversity property, the images obtained from different samplings, \textit{i.e.} ($\tilde{I}_N^1$, $\tilde{I}_V^2$) and ($\tilde{I}_N^2$, $\tilde{I}_V^1$), are regarded as negative pairs.
Note that we do not set ($\tilde{I}_N^1$, $\tilde{I}_N^2$) and ($\tilde{I}_V^1$, $\tilde{I}_V^2$) as negative pairs because HFR is dedicated to cross-domain matching.

Formally, the contrastive loss for the generated data is:
\begin{equation} \label{contrastive}
\begin{split}
    \mathcal{L}_{\text{cont}} = & \sum_{j \neq k}~(1 -  \langle \tilde{f}_N^j, \tilde{f}_V^j \rangle) + (1 - \langle \tilde{f}_N^k, \tilde{f}_V^k \rangle) \\
    & + \text{max}(0, \langle \tilde{f}_N^j, \tilde{f}_V^k \rangle - m) + \text{max}(0, \langle \tilde{f}_N^k, \tilde{f}_V^j \rangle - m),
\end{split}
\end{equation}
where $\tilde{f}_N^j = F(\tilde{I}_N^j)$, $\tilde{f}_V^j = F(\tilde{I}_V^j)$, and $m$ is a margin value. 
By minimizing Eq.~(\ref{contrastive}), the first two terms assist in reducing domain discrepancy, while the last two terms facilitate the learning of discriminative embedding features.
Considering both the softmax loss $\mathcal{L}_{\text{cls}}$ and the contrastive loss $\mathcal{L}_{\text{cont}}$, the overall loss for the HFR network is given by:
\begin{equation} \label{hfr}
    \mathcal{L}_{\text{hfr}} = \mathcal{L}_{\text{cls}} + \alpha \mathcal{L}_{\text{cont}},
\end{equation}
where $\alpha$ is a trade-off parameter.

\begin{figure}[t]
\centering
\includegraphics[width=0.46\textwidth]{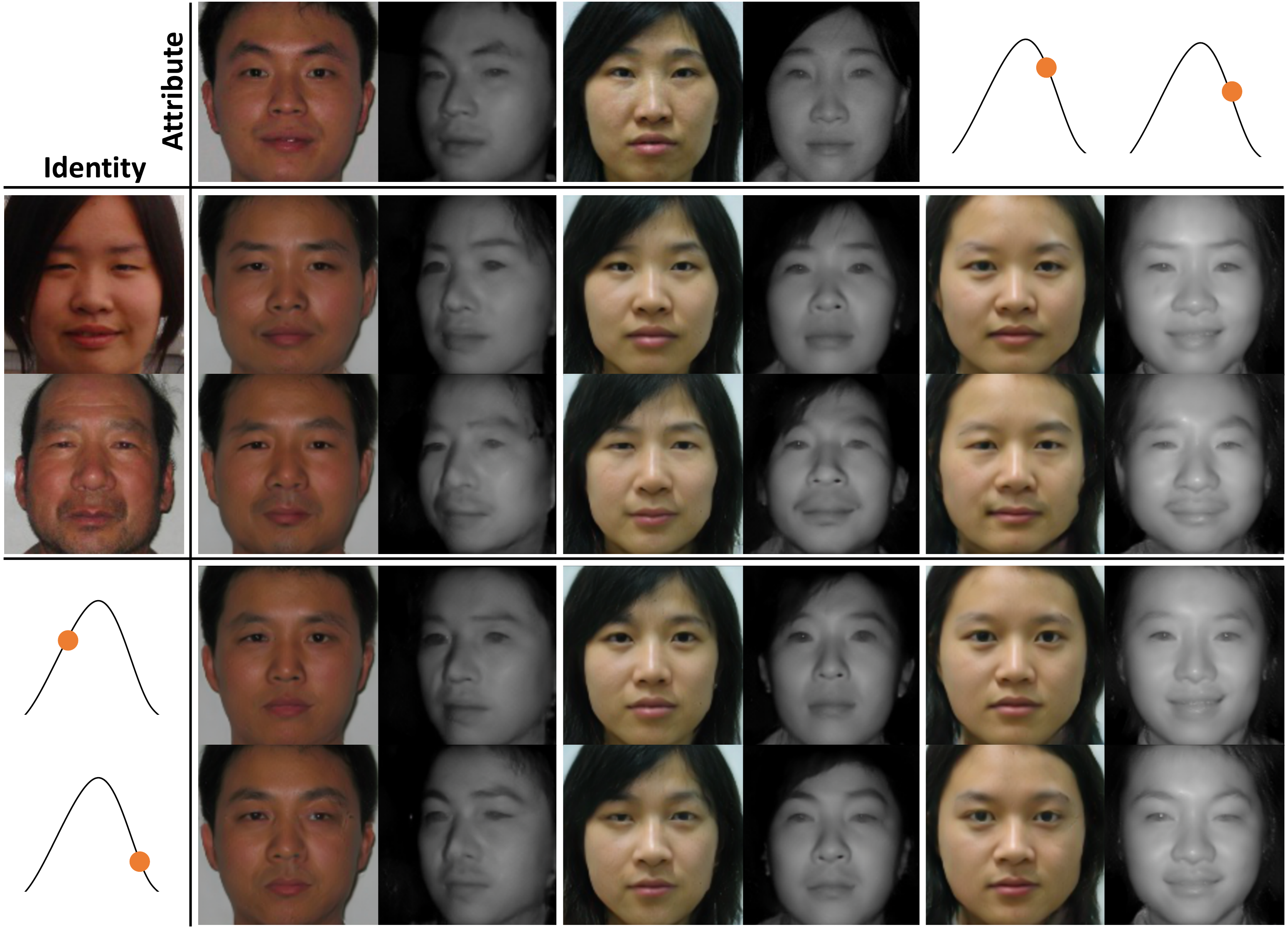}
\caption{Disentanglement experiments on CASIA NIR-VIS 2.0. 
The face images in the first row and the first column are real images, while the remaining face images are generated. After the disentanglement, our method can transfer the image attributes or the sampled attributes (the last two columns) to any identity, including the identities of the real face images and the identities sampled through the identity sampler (the last two rows).
}
\label{fig_3}
\end{figure}

\section{Experiments}
In this section, we report insightful analyses of our proposed method and its evaluations against state-of-the-art methods on seven databases belonging to five HFR tasks, including NIR-VIS, Sketch-Photo, Profile-Frontal Photo, Thermal-VIS, and ID-Camera.
\textbf{Note that the experiments on the Oulu-CASIA NIR-VIS \cite{JChen:2009}, the Multi-PIE \cite{gross2010multi}, and the NJU-ID \cite{huo2017heterogeneous} databases are presented in Sections 1.2.1, 1.2.2, and 1.2.3 of the supplementary material, respectively.}
In the following four subsections, we describe (1) an introduction to the databases and protocols, (2) detailed experimental settings, (3) thorough qualitative and quantitative experimental analyses, and (4) comprehensive comparisons with state-of-the-art methods, respectively.

\subsection{Databases and Protocols}
\emph{CASIA NIR-VIS 2.0 Face Database} \cite{SLi:2013} contains a total of 725 identities, and each identity has 1-22 VIS and 5-50 NIR images.
The protocol consists of ten-fold experiments, each including about 6,100 NIR images and 2,500 VIS images from 360 identities for training. 
Meanwhile, for each fold, 358 VIS images from 358 identities (each identity has one VIS image) constitute the gallery set. Over 6,000 NIR images from the same 358 identities compose the probe set.
There is no identity overlap between the training and the testing.
We conduct experiments on each fold separately, and then report the mean and standard deviation of the results.
The Rank-1 accuracy, Verification Rate (VR) @ False Accept Rate (FAR)=0.1$\%$, and VR@FAR=0.01$\%$ are reported.

\emph{BUAA-VisNir Face Database} \cite{DHuang:2012} is a popular heterogeneous face database that is usually utilized to evaluate domain adaption.
It consists of 150 identities, and each identity includes multiple NIR and VIS images.
The training set has about 1,200 images from 50 identities, and the testing set contains about 1,300 images from the remaining 100 identities.
For testing, only one VIS image per identity is selected as the gallery set and the NIR images constitute the probe set.
We report the Rank-1 accuracy, VR@FAR=1\%, and VR@FAR=0.1$\%$ for comparisons.

\emph{IIIT-D Sketch Viewed Database} \cite{bhatt2012memetic} is developed for Sketch-Photo recognition and contains 238 image pairs.
Among them, 67 pairs are from the FG-NET aging database, 99 pairs are from the Labeled Faces in Wild (LFW) database, and 72 pairs are from the IIIT-D student $\&$ staff database.
Since the IIIT-D Sketch Viewed database only has a small number of images, following \cite{XWu:2018}, we use it as the testing set and adopt CUHK Face Sketch FERET (CUFSF) \cite{ZhangWT11} as the training set. 
CUFSF is also a Sketch-Photo face database that contains 1,194 image pairs.
The Rank-1 accuracy is reported.

\emph{Tufts Face Database} \cite{panetta2018comprehensive} is collected for the study of imaging systems and provides Thermal-VIS face images.
There are a total of 113 identities with 74 females and 39 males from more than 15 countries. 
Each identity has multiple cross-domain face images with different poses.
We randomly select 50 identities as the training set, including 1,490 VIS images and 450 thermal images.
The remaining 63 identities are used as the testing set with 1,916 VIS images and 530 thermal images.
We report the Rank-1 accuracy, VR@FAR=1$\%$, and VR@FAR=0.1$\%$.

The testing process for the HFR network is as follows: Taking the Tufts Face database as an example, the 530 thermal images and the 1,916 VIS images constitute the probe set and the gallery set, respectively. During the testing phase, we first employ the trained HFR network to extract the embedding features of the images in both the probe and the gallery sets. Then, each thermal feature is matched against each VIS feature via calculating the cosine similarity between them, yielding a similarity matrix of size 530$\times$1916. Finally, we calculate the Rank-1 accuracy and the verification rate according to the similarity matrix.

\begin{table}[t]
    \centering
    \caption{Network architectures of the encoder~$E_N / E_V$ and the decoder~$G$. K/S/P denotes kernel size/stride/padding.} 
    \label{table-arc}
    \begin{spacing}{1.2}
    \subfloat[][Encoder architecture.]{
        \label{table-arc-1}
        \resizebox{0.23\textwidth}{!}{
        \begin{tabular}{|ccc|}
            \hline
            LAYER & K/S/P & OUTPUT  \\
            \hline
            Conv1 & 5/1/2 & ~~32 ~$\times$ 128$^2$  \\
            Conv2 & 3/2/1 & ~64 ~$\times$ 64$^2$ \\
            Conv3 & 3/2/1 & ~128 $\times$ 32$^2$ \\
            Conv4 & 3/2/1 & ~256 $\times$ 16$^2$ \\
            Conv5 & 3/2/1 & 256 $\times$ 8$^2$ \\
            Conv6 & 3/2/1 & 256 $\times$ 4$^2$ \\
            \hline
            FC1 & - & 512 \\
            \hline
        \end{tabular}}
    }
    \subfloat[][Decoder architecture.]{
        \label{table-arc-2}
        \resizebox{0.24\textwidth}{!}{
        \begin{tabular}{|ccl|}
            \hline
            LAYER & K/S/P & OUTPUT  \\
            \hline
            FC2 & - & ~~~~2048 \\
            Deconv1 & 2/2/0 & ~128 $\times$ 8$^2$ \\
            Deconv2 & 2/2/0 & ~128 $\times$ 16$^2$ \\
            Deconv3 & 2/2/0 & ~~64 ~$\times$ 32$^2$ \\
            Deconv4 & 2/2/0 & ~~64 ~$\times$ 64$^2$ \\
            Deconv5 & 2/2/0 & ~~32 ~$\times$ 128$^2$ \\
            \hline
            Out & 1/1/0 & ~~~3 ~~$\times$ 128$^2$ \\ 
            \hline
        \end{tabular}}
    }
    \end{spacing}
\end{table}

\subsection{Experimental Settings}
For dual generation, we utilize LightCNN \cite{Wu2018ALC} pre-trained on MS-Celeb-1M \cite{guo2016ms} as the feature extractor $F$.
The network architectures of the encoder $E_V / E_N$ and the decoder $G$ are reported in Table~\ref{table-arc}.
`FC' denotes a fully connected layer.
`Conv' contains a convolutional layer, an instance normalization (IN) layer, and a ReLU activation layer.
Compared with `Conv', `DeConv' replaces the convolutional layer with a transposed convolutional layer and adds a resblock block \cite{he2016deep} after ReLU.
In addition, inspired by \cite{karras2019style,choi2019stargan}, we introduce Adaptive Instance Normalization (AdaIN) \cite{huang2017arbitrary} to facilitate the learning of attributes. 
For NIR and VIS, the affine parameters of AdaIN are learned from the attribute representations $z_N$ and $z_V$ via two extra multilayer perceptrons \cite{huang2018multimodal}, respectively.
Following \cite{karras2019style}, AdaIN is only applied on the low-level layer to avoid destroying the disentanglement of the attribute and the identity.
As a result, IN in 'DeConv5' is replaced by AdaIN.
Furthermore, for NIR and VIS, the weight parameters of the decoder are shared except for these of `Out', which is a single convolutional layer.
All of the networks are trained via an Adam optimizer with a fixed learning rate of 2e-4. The parameters $\lambda_1$ and $\lambda_2$ in Eq.~(\ref{pair}) are set to 50 and 0.5, respectively. We determine the values of the two parameters by balancing the magnitude of the corresponding losses $\mathcal{L}_{\text{ort}}$ and $\mathcal{L}_{\text{ip}}$.

For HFR, we adopt the pre-trained LightCNN as the backbone. Face images are first aligned to a resolution of 144$\times$144, and then are randomly cropped to a resolution of 128$\times$128 as inputs.
One batch consists of 64 real heterogeneous images that are randomly sampled from the real training set, and 64 generated paired heterogeneous images that are randomly sampled from the generated dataset.
We adopt Stochastic Gradient Descent (SGD) as the optimizer, where the learning rate is set to 1e-3 initially and is reduced to 5e-4 gradually. The momentum is set to 0.9 and the weight decay is set to 2e-4. 
The trade-off parameter $\alpha$ and the margin $m$ in Eq.~(\ref{hfr}) are set to 1e-3 and 0.5, respectively.

\begin{table}[t]
    \centering
    \caption{Experimental analyses on Tufts Face. 
    MS (Mean Similarity, higher is better) denotes the mean cosine similarity between the generated paired images. 
    MIS (Mean Instance Similarity, lower is better) presents the mean cosine similarity between two randomly sampled images, including VIS-VIS and NIR-VIS. 
    FID (Fr$\acute{\text{e}}$chet Inception Distance, lower is better) measures the distribution distance between the generated data and the real data. 
    VR@FAR=1$\%$ suggests the recognition performance after using the generated data to train the HFR network via contrastive learning.
    }
    \label{table-1}
    \begin{spacing}{1.3}
    \resizebox{0.49\textwidth}{!}{
    \begin{tabular}{l|ccccc}
        Method & MS & MIS$_\text{VIS-VIS}$ & MIS$_\text{NIR-VIS}$ & FID & VR@FAR=1$\%$ \\
        \myhline
        CoGAN \cite{liu2016coupled} & 0.23 & 0.32 & 0.16 & 0.77 & 22.1 \\
        DVG \cite{fu2019dual}       & 0.52 & 0.45 & 0.41 & 0.65 & 39.3 \\
        \hline
        DVG-Face                    & \textbf{0.53} & \textbf{0.14} & \textbf{0.14} & \textbf{0.57} & \textbf{68.5} \\
    \end{tabular}}
    \end{spacing}
\end{table}

\subsection{Experimental Analyses}
\subsubsection{Identity Consistency} \label{identity consistency}
As discussed in Section~\ref{method}, the generated paired heterogeneous images are required to have the same identity.
In order to verify this property, we propose to measure MS (Mean Similarity) between the paired images.
Specifically, we utilize the pre-trained LightCNN to extract the $L_2$ normalized embedding features of the paired images, and then calculate their cosine similarity.
The MS value is a mean of the similarities of 50K randomly generated paired heterogeneous images.
As reported in Table~\ref{table-1}, the MS value of DVG-Face is 0.53, while that of the real database is 0.31.  
It is observed that the former is even larger than the latter, demonstrating that our method does in fact guarantee the identity consistency property.
In addition, we can see that the preliminary version DVG also has this property due to the usage of the same pairwise identity preserving manner.
By contrast, the MS value of CoGAN \cite{liu2016coupled} is only 0.23 that is significantly smaller than that of the real database. 
This reveals that it is difficult for the weight-sharing manner of CoGAN to ensure the identity consistency property.

\subsubsection{Identity Diversity} \label{identity diversity}
Identity diversity plays a crucial role in our method. The more diverse the generated data, the more valuable it will be for the HFR network. 
Inspired by the instance discrimination criterion \cite{wu2018unsupervised,bachman2019learning} that regards each image as an independent class, we propose a MIS (Mean Instance Similarity) metric to measure the identity diversity property.
Intuitively, if the generated images have abundant identity diversity, the images obtained from different samplings can be considered to have different identities.

Specifically, by randomly sampling twice from noises, two paired images ($\tilde{I}_N^1$, $\tilde{I}_V^1$) and ($\tilde{I}_N^2$, $\tilde{I}_V^2$) are generated. 
Then, we calculate the cosine similarities of the embedding features (extracted by LightCNN) between ($\tilde{I}_V^1$, $\tilde{I}_V^2$) and between ($\tilde{I}_N^1$, $\tilde{I}_V^2$), yielding IS$_\text{VIS-VIS}$ (Instance Similarity) and IS$_\text{NIR-VIS}$, respectively.
After repeating the above sampling and calculation process 50K times, we average these IS$_\text{VIS-VIS}$ and IS$_\text{NIR-VIS}$ values to obtain MIS$_\text{VIS-VIS}$ and MIS$_\text{NIR-VIS}$ values, respectively.
As listed in Table~\ref{table-1}, both CoGAN and DVG get a much higher MIS$_\text{VIS-VIS}$ value than DVG-Face, suggesting the presence of massive similar VIS images in the sampled data of CoGAN and DVG.
We also observe that CoGAN gets a low MIS$_\text{NIR-VIS}$ value due to the identity inconsistency between the generated paired heterogeneous images.
The lowest MIS$_\text{VIS-VIS}$ and MIS$_\text{NIR-VIS}$ values of DVG-Face prove that it can really generate more diverse data.
In addition, when using the generated data to train the HFR network via the contrastive loss in Eq.~(\ref{contrastive}), it is observed that both CoGAN and DVG obtain a much worse recognition performance than DVG-Face.
This is because that the validity of the contrastive loss is based on the identity consistency and the identity diversity properties, neither of which is simultaneously satisfied by CoGAN and DVG.

\begin{figure}[t]
\centering
\includegraphics[width=0.48\textwidth]{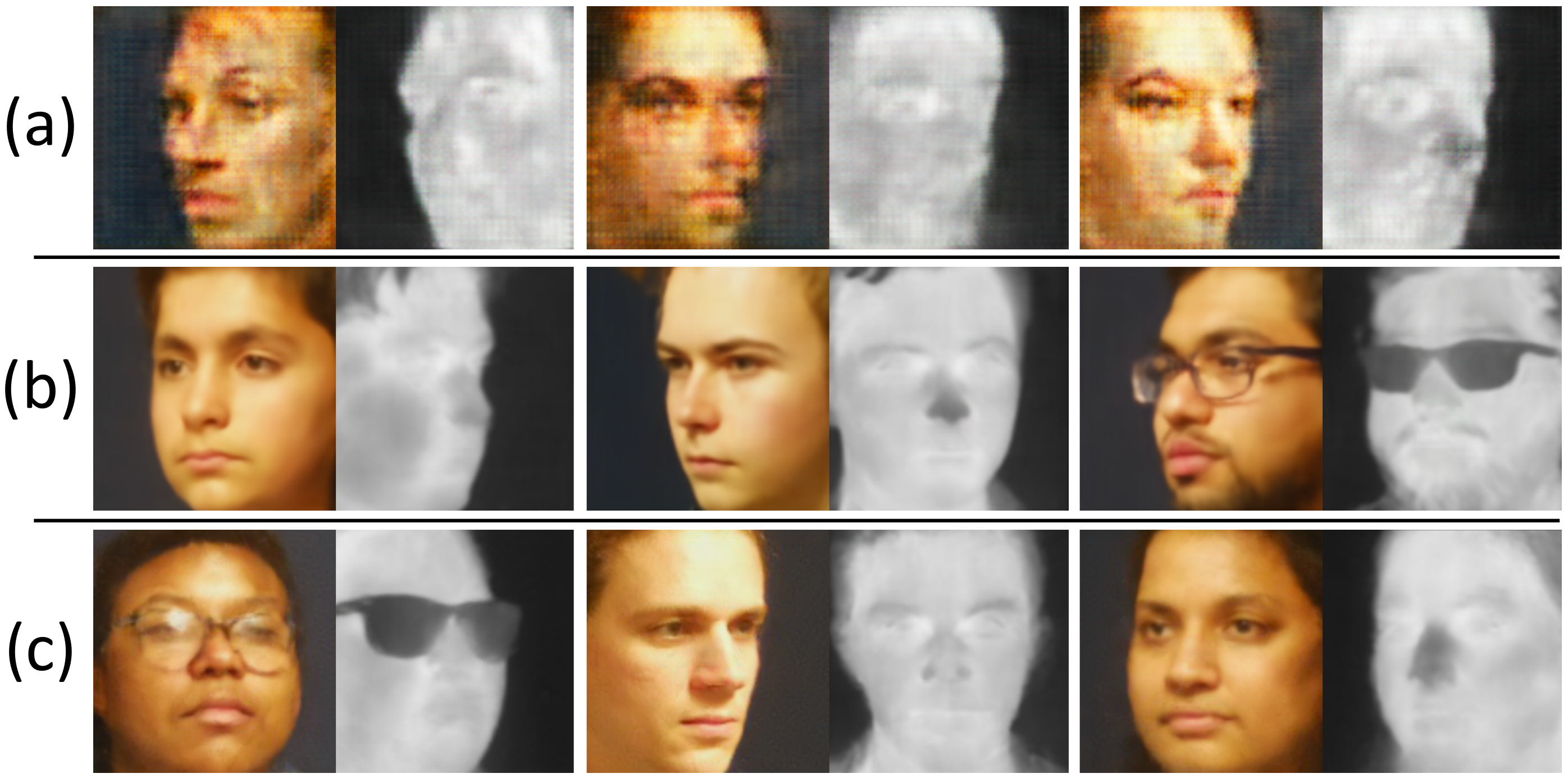}
\caption{Visualization comparisons of dual generation on Tufts Face. (a) CoGAN \cite{liu2016coupled}. (b) DVG \cite{fu2019dual}. (c) DVG-Face.
}
\label{fig_5}
\end{figure}

\begin{table}[t]
    \centering
    \caption{Recognition performances under different identity sampling approaches on Tufts Face. 
    From top to bottom, the first one leverages LightCNN in both the training and the testing stages of dual generation. 
    The second one uses LightCNN in the training stage and the identity sampler in the testing stage. 
    The third one employs the identity sampler in the both stages.}
    \label{table-2}
    \begin{spacing}{1.3}
    \resizebox{0.49\textwidth}{!}{
    \begin{tabular}{l|ccc}
        Method & Rank-1 & VR@FAR=1$\%$ & VR@FAR=0.1$\%$ \\
        \myhline
        LightCNN - LightCNN & 74.6 & 67.9 & 33.7 \\
        LigthCNN - Identity Sampler & 73.5 & 66.4 & 31.0 \\
        Identity Sampler - Identity Sampler & \textbf{75.7} & \textbf{68.5} & \textbf{36.5} \\
    \end{tabular}}
    \end{spacing}
\end{table}

\subsubsection{Distribution Consistency}
The distribution of the generated data and that of the real data should be consistent, otherwise it will be difficult to boost the recognition performance with the generated data.
FID (Fr$\acute{\text{e}}$chet Inception Distance) \cite{heusel2017gans} is employed to measure the distance between these two distributions in the feature space. 
In addition, considering the advantage of the face recognition network in extracting facial features, we use LightCNN rather than the conventional Inception model as the feature extractor.
Specifically, we first generate 50K paired NIR-VIS data, and then calculate the FID value between the generated NIR data and the real NIR data, as well as the FID value between the generated VIS data and the real VIS data. Subsequently, these two FID values are averaged to get the final FID value.
Table~\ref{table-1} suggests that both DVG and DVG-Face obtain a much lower FID value than CoGAN, proving that our method can really learn the distribution of the real data.

Furthermore, Fig.~\ref{fig_5} shows the visualization comparisons of CoGAN, DVG, and DVG-Face.
We can see that the generated results of CoGAN have many artifacts. It seems difficult for CoGAN to generate good results under such a large domain discrepancy case. Besides, it is observed that the generated paired images of CoGAN are not particularly similar, which is consistent with the MS value in Table~\ref{table-1}. 
By contrast, both DVG and DVG-Face generate photo-realistic results, further demonstrating the distribution consistency property of our method.

\begin{figure}[t]
\centering
\includegraphics[width=0.49\textwidth]{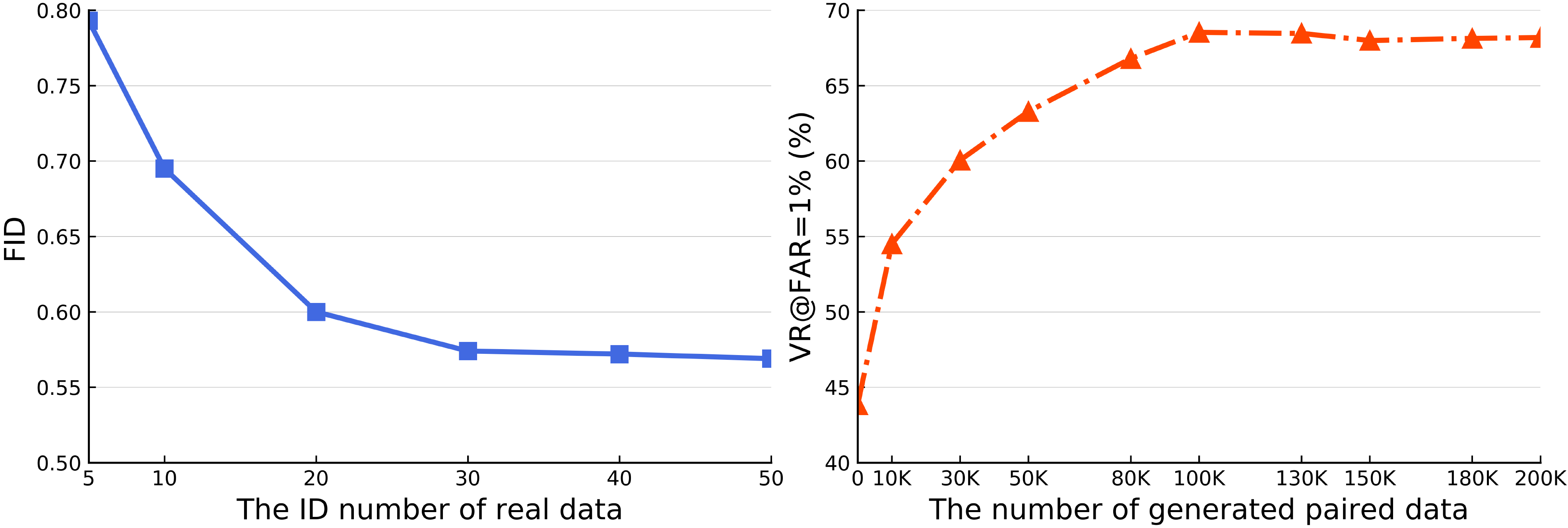}
\caption{Studies of the real data (left) and the generated data (right) on Tufts Face.
}
\label{fig_7}
\end{figure}

In the end, it is obvious that the real training data are crucial for the dual variational generator to learn domain distributions. This motivates us to further investigate the effect of the number of the real data on domain learning.
Specifically, apart from the whole training set of Tufts Face that has 1,940 images belonging to 50 identities, we also train the generator on 5 subsets that include 5, 10, 20, 30, and 40 identities with 198, 378, 734, 1,130, and 1,499 images, respectively. Subsequently, we measure the domain distances between the generated data of the above settings and the real data of the whole training set via FID. The lower the FID value, the closer the domain of the generated data is to that of the real data. The experimental results in Fig.~\ref{fig_7} (left) suggest that as the number of the trained identities increases, the FID value decreases gradually before 30 identities and then tends to be stable. Thus, 30 identities containing 1,130 real training images seem enough for the generator to learn the real domain.

\subsubsection{Identity Sampling} \label{identity sampling}
As stated in Section~\ref{training unpair}, the identity representation can be obtained from the pre-trained face recognition network (LightCNN~\cite{Wu2018ALC}) or from the identity sampler.
Correspondingly, there are three potential identity sampling approaches:
(1) Using LightCNN in both the training and the testing stages of the dual generation. In this case, DVG-Face becomes a conditional generative model. At the testing stage, if we desire to generate large-scale paired data, we must have the same number of VIS data with different identities to provide the required identity representations.
Therefore, the sampling diversity is limited by the size of the available VIS data.
(2) Using LightCNN in the training stage and the identity sampler in the testing stage.
By this means, we no longer need large-scale available VIS data in the testing stage.
The required rich identity representations can be easily sampled from noises.
However, the identity representations generated by the identity sampler conform to the Gaussian distribution, while those of LightCNN do not satisfy the Gaussian distribution hypothesis.
Thus, there may be a distribution gap between the two types of identity representations.
(3) Using the identity sampler in both the training and the testing stages. In this way, it is flexible to obtain massive identity representations in the both stages.

The performances of the above three approaches are shown in Table~\ref{table-2}.
We observe that the second approach obtains the worst results because of the aforementioned distribution gap.
In addition, the third approach gets better results than the first approach, even though the identity sampler is trained on the extracted identity representations of LightCNN. 
This indicates that the identity sampler may produce new diverse identity representations based on the property of VAEs.

\begin{table}[t]
    \centering
    \caption{Results under different usages of large-scale VIS data on Tufts Face. 
    (A) trains LightCNN on the large-scale VIS database MS-Celeb-1M.
    (B) fine-tunes the pre-trained LightCNN on Tufts Face.
    (C) fine-tunes on both Tufts Face and MS-Celeb-1M.
    (D) is our method. It introduces the identities of MS-Celeb-1M to enrich the identity diversity of the generated data, and fine-tunes the pre-trained LightCNN on both Tufts Face and the generated data.
    }
    \label{table-3}
    \begin{spacing}{1.0}
    \resizebox{0.45\textwidth}{!}{
    \begin{tabular}{c|ccc}
        Method & Rank-1 & VR@FAR=1$\%$ & VR@FAR=0.1$\%$ \\
        \myhline
        (A) & 29.4 & 23.0 & 5.3 \\
        (B) & 54.5 & 42.4 & 15.6 \\
        (C)  & 33.9 & 24.7 & 7.9 \\
        \hline
        (D) & \textbf{75.7} & \textbf{68.5} & \textbf{36.5} \\
    \end{tabular}}
    \end{spacing}
\end{table}

\subsubsection{Number of Generated Data}
We also investigate the effect of the number of the generated data on the HFR performance.
In particular, we use 0, 10K, 30K, 50K, 80K, 100K, 130K, 150K, 180K, and 200K generated paired heterogeneous data to train the HFR network, respectively.
Fig.~\ref{fig_7} (right) depicts the results under different numbers of generated data on the Tufts Face database.
It is clear that from 0 to 100K, the recognition performance is improved with increasing the number of the generated data.
Compared with only using the real data (0 generated data), VR@FAR=1$\%$ is significantly improved by 26.1$\%$ after adding 100K generated paired data.
These phenomena demonstrate that the generated data really contain abundant useful information that is helpful for recognition.
Besides, as the number of the generated paired data exceeds 100K, no obvious improvement is observed, indicating that the generated data have been saturated.

\begin{figure}[t]
\centering
\includegraphics[width=0.49\textwidth]{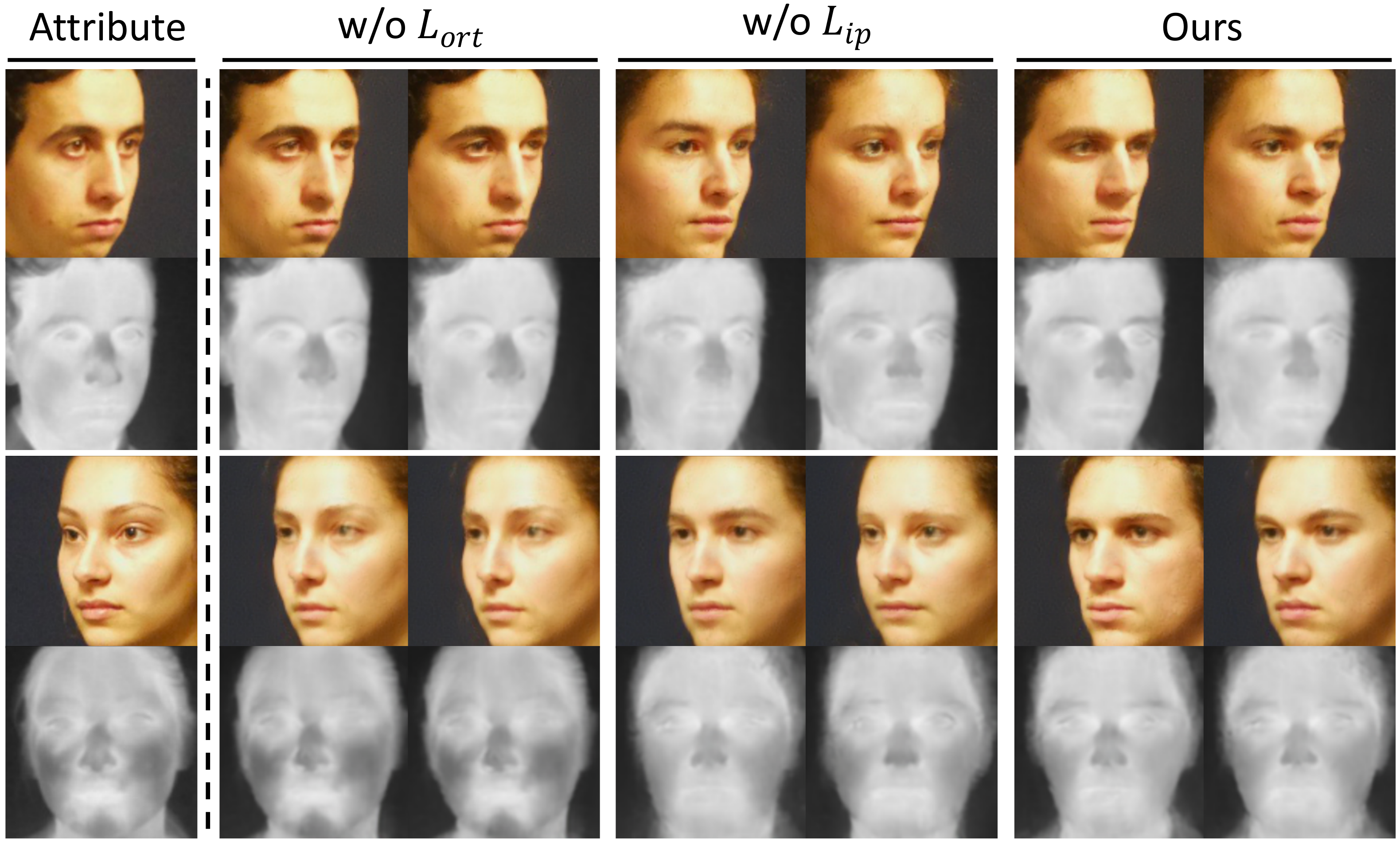}
\caption{Qualitative ablation studies of the angular orthogonal loss $\mathcal{L}_{\text{ort}}$ and the pairwise identity preserving loss $\mathcal{L}_{\text{ip}}$ on Tufts Face. Given the attribute images, these methods generate two pairs of images with two different identity representations.
}
\label{fig_6}
\end{figure}

\begin{table}[t]
    \centering
    \caption{Quantitative ablation studies of the angular orthogonal loss $\mathcal{L}_{\text{ort}}$ and the pairwise identity preserving loss $\mathcal{L}_{\text{ip}}$ on Tufts Face.}
    \label{table-5}
    \begin{spacing}{1.1}
    \resizebox{0.43\textwidth}{!}{
    \begin{tabular}{c|ccc}
        Method & Rank-1 & VR@FAR=1$\%$ & VR@FAR=0.1$\%$ \\
        \myhlineII
        w/o $\mathcal{L}_{\text{ort}}$ & 55.9 & 43.6 & 16.7 \\
        w/o $\mathcal{L}_{\text{ip}}$ & 62.4 & 57.6 & 22.1 \\
        \hline
        Ours & \textbf{75.7} & \textbf{68.5} & \textbf{36.5} \\
    \end{tabular}}
    \end{spacing}
\end{table}

\subsubsection{Usage of Large-Scale VIS Data} \label{VIS data}
Generally, the large-scale VIS database, \textit{i.e.} MS-Celeb-1M \cite{guo2016ms}, is only used to pre-train the HFR network \cite{deng2019residual,XWu:2018}.
The intuition behind this is that training the large-scale VIS data together with the small-scale heterogeneous data will lead to severe domain biases that may degrade the performance of HFR.
In order to verify this point, four experiments are elaborately designed:
(A) directly trains LightCNN on the large-scale VIS database MS-Celeb-1M.
(B) fine-tunes the pre-trained LightCNN on the heterogeneous database, and (C) fine-tunes on both the heterogeneous database and MS-Celeb-1M.
(D) is our method that fine-tunes on both the heterogeneous database and the generated data.

Table~\ref{table-3} shows that the results of (C) are merely slightly better than those of (A), and are much worse than those of (B), revealing that the domain biases actually affect the performance of HFR.
Moreover, (D) outperforms the others by a large margin. This demonstrates the effectiveness of our method that introduces the identity representations of MS-Celeb-1M to enrich the identity diversity of the generated data.
In this case, the domain biases are mitigated because the generated data belong to the same domain of the real heterogeneous database.

\begin{table}[t]
    \centering
    \caption{Results under different usages of the generated data on Tufts Face.
    (I) only uses the real data to train the HFR network, and (II) merely uses the generated unlabeled data.
    (III) leverages both the real data and the generated data, and the latter is utilized via the pairwise distance loss.
    (IV) is our method that replaces the pairwise distance loss with the contrastive loss.
    }
    \label{table-5-2}
    \begin{spacing}{1.0}
    \resizebox{0.45\textwidth}{!}{
    \begin{tabular}{c|ccc}
        Method & Rank-1 & VR@FAR=1$\%$ & VR@FAR=0.1$\%$ \\
        \myhline
        (I) & 54.5 & 42.4 & 15.6 \\
        (II) & 36.6 & 24.7 & 7.2 \\
        (III) & 70.3 & 60.8 & 29.6 \\
        \hline
        (IV) & \textbf{75.7} & \textbf{68.5} & \textbf{36.5} \\
    \end{tabular}}
    \end{spacing}
\end{table}

\begin{table}[t]
    \centering
    \caption{Verification rates on Tufts Face under different parameter values.} 
    \label{table-6}
    \begin{spacing}{0.9}
    \subfloat[][$\lambda_1$ in Eq.~(\ref{pair})]{
        \label{table-6-1}
        \resizebox{0.115\textwidth}{!}{
        \begin{tabular}{cc}
            $\lambda_1$ & FAR=1$\%$  \\
            \toprule[0.95 pt]
            0 & 43.6 \\
            10 & 68.1 \\
            25 & \textbf{68.5} \\
            50 & \textbf{68.5} \\
            75 & 67.9 \\
            100 & 67.4 \\
        \end{tabular}}
    }
    \subfloat[][$\lambda_2$ in Eq.~(\ref{pair})]{
        \label{table-6-2}
        \resizebox{0.115\textwidth}{!}{
        \begin{tabular}{cc}
            $\lambda_2$ & FAR=1$\%$  \\
            \toprule[0.95 pt]
            0 & 57.6 \\
            0.10 & 66.9 \\
            0.25 & 67.8 \\
            0.50 & \textbf{68.5} \\
            0.75 & 68.4 \\
            1.00 & 68.3 \\
        \end{tabular}}
    }
    \subfloat[][$\eta$ in Eq.~(\ref{rec-u})]{
        \label{table-6-4}
        \resizebox{0.115\textwidth}{!}{
        \arrayrulecolor{black}
        \begin{tabular}{cc}
            $\eta$ & FAR=1$\%$  \\
            \toprule[0.95 pt]
            0    & 46.5 \\
            0.05 & 68.1 \\
            0.10 & \textbf{68.5} \\
            0.20 & 67.6 \\
            0.30 & 65.7 \\
            0.40 & 64.8 \\
        \end{tabular}}
    }
    \subfloat[][$\alpha$ in Eq.~(\ref{hfr})]{
        \label{table-6-3}
        \resizebox{0.115\textwidth}{!}{
        \arrayrulecolor{black}
        \begin{tabular}{cc}
            $\alpha$ & FAR=1$\%$  \\
            \toprule[0.95 pt]
            0 & 42.4 \\
            2e-4 & 67.9 \\
            5e-4 & 68.3 \\
            1e-3 & \textbf{68.5} \\
            2e-3 & 67.6 \\
            3e-3 & 65.7 \\
        \end{tabular}}
    }
    \end{spacing}
\end{table}

\begin{table}[t]
    \centering
    \caption{Studies of the margin $m$ in Eq.~(\ref{contrastive}) on Tufts Face.}
    \label{table-7}
    \begin{spacing}{1.1}
    \resizebox{0.45\textwidth}{!}{
    \begin{tabular}{c|cccccc}
        margin $m$ & 0.2 & 0.3 & 0.4 & 0.5 & 0.6 & 0.7 \\
        \myhlineII
        VR@FAR=1$\%$ & 67.0 & 67.5 & 68.1 & \textbf{68.5} & 68.2 & 67.4 \\
    \end{tabular}}
    \end{spacing}
\end{table}

\subsubsection{Ablation Study}
\textbf{For Dual Generation.}
The angular orthogonal loss $\mathcal{L}_{\text{ort}}$ in Eq.~(\ref{orthogonal}) and the pairwise identity preserving loss $\mathcal{L}_{\text{ip}}$ in Eq.~(\ref{ip}) (including $\mathcal{L}_{\text{ip}}^{\text{u}}$ in Eq.~(\ref{ip-u})) are studied to analyze their respective roles. 
Both quantitative and qualitative results are presented for a comprehensive comparison.

Fig.~\ref{fig_6} depicts the qualitative comparisons between our method and its two variants. 
Without $\mathcal{L}_{\text{ort}}$, the generated images are almost unchanged when the identity representation is varied.
This means that we cannot enrich the identity diversity of the generated data via injecting more identity information.
When omitting $\mathcal{L}_{\text{ip}}$, the generated paired images look inconsistent in identities, suggesting the effectiveness of $\mathcal{L}_{\text{ip}}$. 
In addition, Table~\ref{table-5} displays the quantitative comparison results.
It is observed that the recognition performance drops significantly if one of the losses is discarded, which further proves the importance of each loss.
For example, VR@FAR=0.1$\%$ decreases by 19.8$\%$ when $\mathcal{L}_{\text{ort}}$ is omitted.

\textbf{For HFR.}
We design four experiments to study the usage of the generated data:
(I) only uses the real data to train the HFR network.
(II) discards the real data and merely uses the generated data.
(III) leverages both the real data and the generated data, and the latter is utilized via the pairwise distance loss as our preliminary version DVG.
(IV) is our method that uses the real data and the generated data with the softmax loss and the contrastive loss, respectively.

Table~\ref{table-5-2} shows that (II) gets the worst results.
However, compared with the backbone LightCNN pre-trained on MS-Celeb-1M (Rank-1 = 29.4$\%$, listed in Table~\ref{table-3}), (II) still gains an improvement of 7.2$\%$ in terms of the Rank-1 accuracy, suggesting the value of the generated data.
Compared with (I), the Rank-1 accuracy is boosted at least by 15.8$\%$ after introducing the generated data, including both (III) and (IV). This further demonstrates the importance of the generated data.
In addition, we find that compared with our preliminary version DVG (Rank-1 = 56.1$\%$, listed in Table~\ref{table-tufts}), (III) improves the Rank-1 accuracy by 14.2$\%$.
Since the difference between the two methods lies in the generated data, we attribute the improvement to the richer identity diversity of the generated data of DVG-Face, as discussed in Section~\ref{identity diversity}.
(IV) gains better results than (III), which verifies the effectiveness of the introduced contrastive loss.
It optimizes both the intra-class and the inter-class distances, facilitating to learn domain-invariant and discriminative embedding features.
By contrast, the pairwise distance loss of DVG can only assist in reducing domain discrepancy.
Note that the effectiveness of the contrastive loss is based on the identity consistency and the identity diversity properties of the generated data, as stated in Section~\ref{identity diversity}.

Table~\ref{table-6} lists the results of sensitivity studies of the parameters $\lambda_1$ and $\lambda_2$ in Eq.~(\ref{pair}), $\eta$ in Eq.~(\ref{rec-u}), and $\alpha$ in Eq.~(\ref{hfr}).
The parameters $\lambda_1$, $\lambda_2$, and $\alpha$ are set to balance the magnitude of the corresponding loss functions.
$\eta$ is set according to \cite{bao2018towards}.
It is observed that our method is not sensitive to these trade-off parameters in a large range. 
For instance, when $\lambda_1$ is varied from 10 to 100, the verification rate only changes from 67.4$\%$ to 68.5$\%$.
Besides, Table~\ref{table-6} suggests that as the parameter becomes larger gradually, the recognition performance first increases and then decreases. The optimal setting is $\lambda_1$=50, $\lambda_2$=0.5, $\eta$=0.1, and $\alpha$=1e-3. More sensitivity studies of the parameters $\lambda_1$, $\lambda_2$, $\eta$, and $\alpha$ on the BUAA-Vis-Nir, the MultiPIE, and the NJU-ID databases are presented in Tables 4, 5, and 6 of the supplementary material, respectively.
Finally, we further explore the setting of the margin parameter $m$ in Eq.~(\ref{contrastive}). 
Table~\ref{table-7} reveals that the best margin value is 0.5.

Finally, it is worth noting that our method is computationally efficient. When using one TITAN Xp GPU, it takes about 20 minutes to train the generator on the Tufts Face database. In the inference stage of the generator, it merely takes about 4.7 ms to generate a pair of heterogeneous face images. Furthermore, training the HFR network with 100K generated data and about 2K real data costs about 5 hours.

\begin{figure*}[t]
\centering
\includegraphics[width=0.98\textwidth]{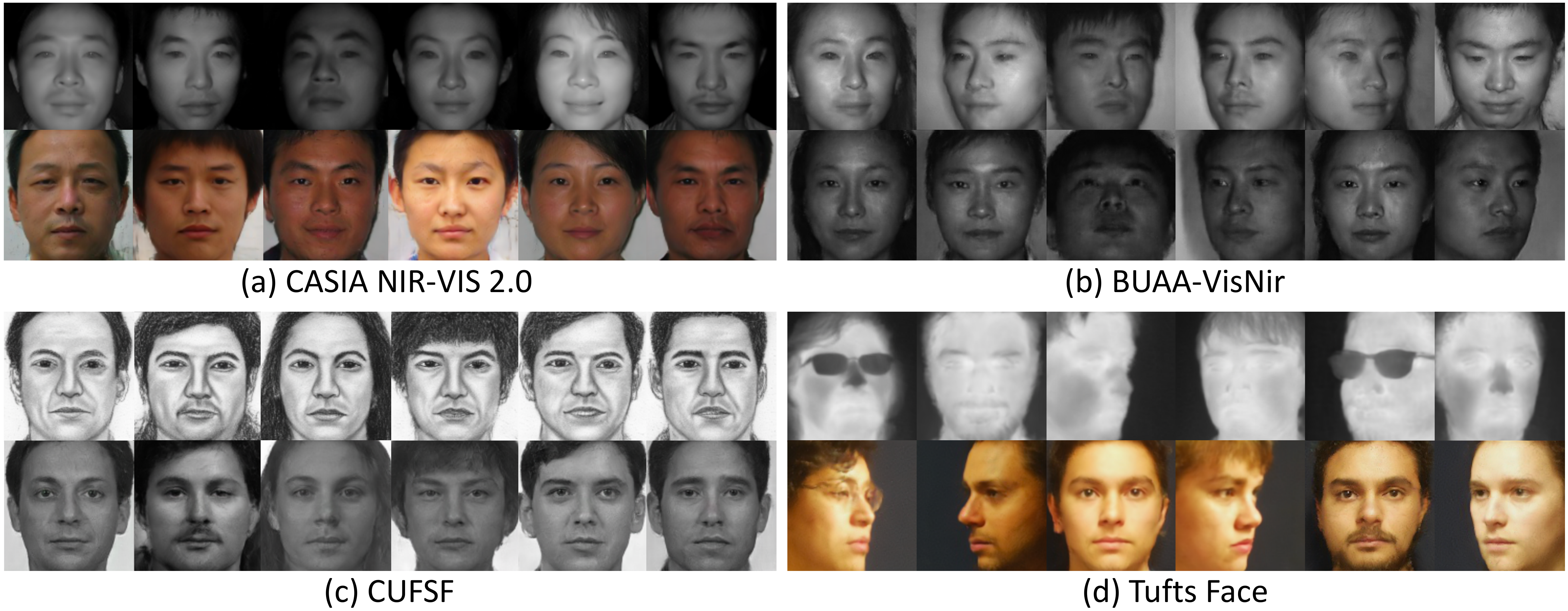}
\caption{Dual generation results on (a) CASIA NIR-VIS 2.0 \cite{SLi:2013}, (b) BUAA-VisNir \cite{DHuang:2012}, (c) CUFSF \cite{ZhangWT11}, and (d) Tufts Face \cite{panetta2018comprehensive}. More generation results, including those on Oulu-CASIA NIR-VIS \cite{JChen:2009} and Multi-PIE \cite{gross2010multi}, are shown in Fig.~2 of the supplementary material.
}
\label{fig_4}
\end{figure*}

\begin{table}[t]
    \centering
    \caption{Experimental results on CASIA NIR-VIS 2.0.}
    \label{table-casia}
    \begin{spacing}{1.1}
    \resizebox{0.47\textwidth}{!}{
    \begin{tabular}{l|ccc}
        Method & Rank-1 & VR@FAR=0.1$\%$ & VR@FAR=0.01$\%$ \\
        \myhline
        H2(LBP3) \cite{Shao2017CrossModalityFL} & 43.8 & 10.1 & - \\
        DSIFT \cite{dhamecha2014effectiveness} & 73.3$\pm$1.1 & - & - \\
        CDFL \cite{jin2015coupled} & 71.5$\pm$1.4 & 55.1 & - \\
        Gabor+RBM \cite{yi2015shared} & 86.2$\pm$1.0 & 81.3$\pm$1.8 & -  \\
        Recon.+UDP \cite{JuefeiXu2015NIRVISHF} & 78.5$\pm$1.7 & 85.8 & - \\
        CEFD \cite{gong2017heterogeneous} & 85.6 & - & - \\
        \hline
        IDNet \cite{Reale2016SeeingTF} & 87.1$\pm$0.9 & 74.5 & - \\
        HFR-CNN \cite{saxena2016heterogeneous} & 85.9$\pm$0.9 & 78.0 & - \\
        Hallucination \cite{Lezama2017NotAO} & 89.6$\pm$0.9 & - & - \\
        DLFace \cite{peng2019dlface} & 98.68 & - & - \\
        TRIVET \cite{XXLiu:2016} & 95.7$\pm$0.5 & 91.0$\pm$1.3 & 74.5$\pm$0.7 \\
        W-CNN \cite{DBLP:journals/corr/abs-1708-02412} & 98.7$\pm$0.3 & 98.4$\pm$0.4 & 94.3$\pm$0.4 \\
        PACH \cite{duan2019pose} & 98.9$\pm$0.2 & 98.3$\pm$0.2 & - \\
        RCN \cite{deng2019residual} & 99.3$\pm$0.2 & 98.7$\pm$0.2 & - \\
        MC-CNN \cite{8624555} & 99.4$\pm$0.1 & 99.3$\pm$0.1 & - \\
        DVR \cite{XWu:2019}  & 99.7$\pm$0.1 & 99.6$\pm$0.3 & 98.6$\pm$0.3 \\
        \hline
        DVG \cite{fu2019dual} & 99.8$\pm$0.1 & 99.8$\pm$0.1 & 98.8$\pm$0.2 \\
        \textbf{DVG-Face} & \textbf{99.9}$\pm$0.1 & \textbf{99.9}$\pm$0.0 & \textbf{99.2}$\pm$0.1 \\
    \end{tabular}}
    \end{spacing}
\end{table}

\begin{figure}[t]
\centering
\includegraphics[width=0.485\textwidth]{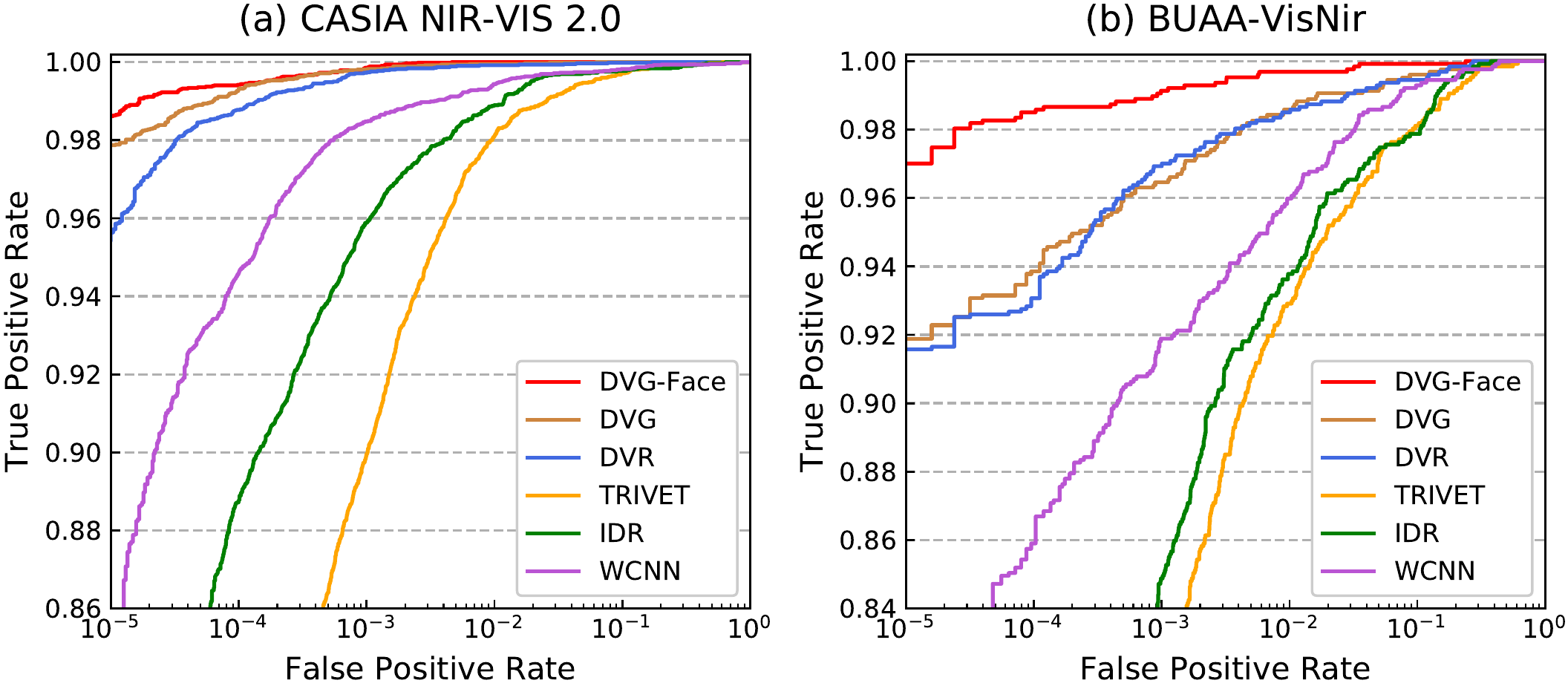}
\caption{ROC curves of different methods on (a) CASIA NIR-VIS 2.0 and (b) BUAA-VisNir.
}
\label{fig_roc}
\end{figure}

\subsection{Comparisons with State-of-the-Art Methods}
\subsubsection{Results on the CASIA NIR-VIS 2.0 Database}
In order to verify the effectiveness of DVG-Face, we compare it with 16 state-of-the-art methods, including 6 traditional methods and 10 deep learning based methods.
The traditional methods include H2(LBP3) \cite{Shao2017CrossModalityFL}, DSIFT+PCA+LDA \cite{dhamecha2014effectiveness}, CDFL \cite{jin2015coupled}, Gabor+RBM \cite{yi2015shared}, Recon.+UDP \cite{JuefeiXu2015NIRVISHF}, and CEFD \cite{gong2017heterogeneous}.
The deep learning based methods consist of IDNet \cite{Reale2016SeeingTF}, HFR-CNN \cite{saxena2016heterogeneous}, Hallucination \cite{Lezama2017NotAO}, DLFace \cite{peng2019dlface}, TRIVET \cite{XXLiu:2016}, W-CNN \cite{DBLP:journals/corr/abs-1708-02412}, PACH \cite{duan2019pose}, RCN \cite{deng2019residual}, MC-CNN \cite{8624555}, and DVR \cite{XWu:2019}.

Table~\ref{table-casia} shows the results of the Rank-1 accuracy and verification rates of the above methods. 
We can see that most of the deep learning based methods outperform the traditional methods. 
Nevertheless, the deep learning based method HFR-CNN performs worse than the traditional method Gabor+RBM. 
This indicates that it is challenging for the deep learning based methods to train on the small-scale heterogeneous data.
The methods, such as W-CNN, MC-CNN, and DVR, which are devoted to tackling the over-fitting caused by the limited training data, get higher performances.
Moreover, our method obtains the best results than all the competitors.
In particular, regarding the Rank-1 accuracy and VR@FAR=0.1$\%$, DVG-Face exceeds the conditional generation based method PACH by 1.0$\%$ and 1.6$\%$ respectively, suggesting the superiority of our unconditional generation approach.
Compared with the state-of-the-art method DVR, we improve VR@FAR=0.01$\%$ from 98.6$\%$ to 99.2$\%$.
More importantly, the ROC curves in Fig.~\ref{fig_roc} (a) show that DVG-Face improves the harder indicator True Positive Rate (TPR)@False Positive Rate (FPR)=10$^{-5}$ by 3.0$\%$ over DVR.
In addition, compared with the preliminary version DVG \cite{fu2019dual}, DVG-Face further improves TPR@FPR=10$^{-5}$ from 97.8$\%$ to 98.6$\%$. The improvement highlights the importance of the abundant identity diversity and the consequent contrastive loss.
The generated paired NIR-VIS images of DVG-Face are exemplified in Fig.~\ref{fig_4} (a).

\begin{table}[t]
    \centering
    \caption{Experimental results on BUAA-VisNir.}
    \label{table-buaa}
    \begin{spacing}{1.1}
    \resizebox{0.47\textwidth}{!}{
    \begin{tabular}{l|ccc}
        Method & Rank-1 & VR@FAR=1$\%$ & VR@FAR=0.1$\%$ \\
        \myhline
        MPL3 \cite{JChen:2009} & 53.2 & 58.1 & 33.3 \\
        KCSR \cite{lei2009coupled} & 81.4 & 83.8 & 66.7 \\
        KPS \cite{klare2012heterogeneous} & 66.6 & 60.2 & 41.7 \\
        KDSR \cite{Huang2013RegularizedDS} & 83.0 & 86.8 & 69.5 \\
		H2(LBP3) \cite{Shao2017CrossModalityFL} & 88.8 & 88.8 & 73.4 \\
		\hline
		TRIVET \cite{XXLiu:2016} & 93.9 & 93.0 & 80.9 \\
        W-CNN \cite{DBLP:journals/corr/abs-1708-02412} & 97.4 & 96.0 & 91.9 \\
        PACH \cite{duan2019pose} & 98.6 & 98.0 & 93.5 \\
        DVR \cite{XWu:2019}  & 99.2 & 98.5 & 96.9 \\
       	\hline
       	DVG \cite{fu2019dual} & 99.3 & 98.5 & 97.3 \\
        \textbf{DVG-Face} & \textbf{99.9} & \textbf{99.7} & \textbf{99.1}
    \end{tabular}}
    \end{spacing}
\end{table}

\subsubsection{Results on the BUAA-VisNir Database}
A total of 5 traditional methods and 4 deep learning based methods are compared on the BUAA-VisNir database.
The former includes MPL3 \cite{JChen:2009}, KCSR \cite{lei2009coupled}, KPS \cite{klare2012heterogeneous}, KDSR \cite{Huang2013RegularizedDS}, and H2(LBP3) \cite{Shao2017CrossModalityFL}, and the latter includes TRIVET~\cite{XXLiu:2016}, W-CNN~\cite{DBLP:journals/corr/abs-1708-02412}, PACH \cite{duan2019pose}, and DVR~\cite{XWu:2019}.

Here as well, we observe from Table~\ref{table-buaa} that the deep learning based methods are always superior to the traditional methods.
Moreover, the improvements of DVG-Face over the conditional generation based method PACH are significant. VR@FAR=0.1$\%$ is increased from 93.5$\%$ to 99.1$\%$.
In addition, although the state-of-the-art DVR has achieved very high results, DVG-Face still exceeds it by 2.2$\%$ in terms of VR@FAR=0.1$\%$.
The ROC curves in Fig.~\ref{fig_roc} (b) further demonstrate the advantages of our method.
Regarding TPR@FPR=10$^{-5}$, DVG-Face leads DVR by 5.4$\%$ and the preliminary version DVG by 5.1$\%$.
The generated paired images of DVG-Face on the BUAA-VisNir database are presented in Fig.~\ref{fig_4} (b).

\subsubsection{Results on the IIIT-D Sketch Viewed Database}
Considering the crucial role of the Sketch-Photo recognition in criminal investigation, we further evaluate our method on the IIIT-D Sketch Viewed database.
The compared methods include 5 traditional methods: Original WLD \cite{chen2009wld}, SIFT \cite{klare2010sketch}, EUCLBP \cite{bhatt2010matching}, LFDA \cite{Klare2011MatchingFS}, MCWLD \cite{bhatt2012memetic}, as well as 4 deep learning based methods: LightCNN \cite{Wu2018ALC}, CDL \cite{XWu:2018}, RCN\cite{deng2019residual}, and MC-CNN~\cite{8624555}.

Table~\ref{table-sketch} suggests that on such a Sketch-Photo database, the overall performance discrepancy between the traditional methods and the deep learning based methods is not particularly large. 
The Rank-1 accuracy of MCWLD is even better than that of LightCNN, and is comparable with that of CDL.
In addition, our method exhibits superior performances over all the competitors.
The best Rank-1 accuracy is improved from 90.34$\%$ \cite{deng2019residual} to 97.21$\%$.
Fig.~\ref{fig_4} (c) depicts the dual generation results of Sketch-Photo.

\begin{table}[t]
    \centering
    \caption{Experimental results on IIIT-D Sketch Viewed.}
    \label{table-sketch}
    \begin{spacing}{1.1}
    \resizebox{0.27\textwidth}{!}{
    \begin{tabular}{l|c}
        Method & Rank-1  \\
        \myhline
        Original WLD \cite{chen2009wld} & 74.34 \\
        SIFT \cite{klare2010sketch} & 76.28 \\
        EUCLBP \cite{bhatt2010matching} & 79.36 \\
        LFDA \cite{Klare2011MatchingFS} & 81.43 \\
        MCWLD \cite{bhatt2012memetic} & 84.24 \\
        \hline
        LightCNN \cite{Wu2018ALC} & 83.24 \\
        CDL \cite{XWu:2018} & 85.35 \\
        MC-CNN \cite{8624555} & 87.40 \\
        RCN \cite{deng2019residual} & 90.34 \\
        \hline
        DVG \cite{fu2019dual} & 96.99 \\
        \textbf{DVG-Face} & \textbf{97.21} \\
    \end{tabular}}
    \end{spacing}
\end{table}

\subsubsection{Results on the Tufts Face Database}
As mentioned in Section~\ref{VIS data}, the face recognition network pre-trained on MS-Celeb-1M merely obtains a Rank-1 accuracy of 29.4$\%$ on the Tufts Face database, indicating the large domain discrepancy between the thermal data and the VIS data.
At the same time, it is observed that after fine-tuning on the training set of Tufts Face, the Rank-1 accuracy only increases to 54.5$\%$ that is still unsatisfactory. This phenomenon is caused by the limited number of the Thermal-VIS training data.
Obviously, data augmentation is a straightforward approach to alleviate this problem.
Both the preliminary version DVG and the current DVG-Face aim to generate more data to facilitate the training of the HFR network.
The comparisons of DVG, DVG-Face, and the baseline LightCNN are reported in Table~\ref{table-tufts}.
We can see that compared with DVG, DVG-Face boosts the Rank-1 accuracy from 56.1$\%$ to 75.7$\%$, and VR@FAR=0.1$\%$ from 17.1$\%$ to 36.5$\%$.
The dual generation results on Tufts Face are shown in Fig.~\ref{fig_4} (d).

\begin{table}[t]
    \centering
    \caption{Experimental results on Tufts Face.}
    \label{table-tufts}
    \begin{spacing}{1.1}
    \resizebox{0.45\textwidth}{!}{
    \begin{tabular}{l|ccc}
        Method & Rank-1 & VR@FAR=1$\%$ & VR@FAR=0.1$\%$  \\
        \myhline
        LightCNN \cite{Wu2018ALC} & 29.4 & 23.0 & 5.3 \\
        \hline
        DVG \cite{fu2019dual} & 56.1 & 44.3 & 17.1 \\
        \textbf{DVG-Face} & \textbf{75.7} & \textbf{68.5} & \textbf{36.5} \\
    \end{tabular}}
    \end{spacing}
\end{table}

\section{Conclusion}
This paper has proposed a novel DVG-Face framework that generates large-scale new paired heterogeneous images from noises to boost the performance of HFR.
To begin with, a dual variational generator is elaborately designed to train with both paired heterogeneous data and unpaired VIS data.
The introduction of the latter greatly promotes the identity diversity of the generated images.
Subsequently, a pairwise identity preserving loss is imposed on the generated paired images to guarantee their identity consistency.
Finally, benefiting from the identity consistency and identity diversity properties, the generated unlabeled images can be employed to train the HFR network via contrastive learning. 
Our method obtains the best results on seven databases, exposing a new way to the HFR problem.

\section*{Acknowledgments}
The authors would like to greatly thank the associate editor and the reviewers for their valuable comments and advice.
This work is funded by Beijing Natural Science Foundation (Grant No. JQ18017), National Natural Science Foundation of China (Grant No. 61721004, U20A20223) and Youth Innovation Promotion Association CAS (Grant No. Y201929).

\ifCLASSOPTIONcaptionsoff
  \newpage
\fi

{
\bibliographystyle{IEEEtran}
\bibliography{mybibfile}
}

\cleardoublepage
%\includepdfmerge{supplementary_material.pdf,1}


\section*{Supplementary material}

# Supplementary Material

---❖---

TABLE 1: Experimental results on Oulu-CASIA NIR-VIS.

| Method | Rank-1 | VR@FAR=1% | VR@FAR=0.1% |
|--------|--------|-----------|-------------|
| MPL3 | 48.9 | 41.9 | 11.4 |
| KCSR | 66.0 | 49.7 | 26.1 |
| KPS | 62.2 | 48.3 | 22.2 |
| KDSR | 66.9 | 56.1 | 31.9 |
| H2(LBP3) | 70.8 | 62.0 | 33.6 |
| TRIVET | 92.2 | 67.9 | 33.6 |
| W-CNN | 98.0 | 81.5 | 54.6 |
| PACH | **100.0** | 97.9 | 88.2 |
| DVR | **100.0** | 97.2 | 84.9 |
| DVG | **100.0** | 98.5 | 92.9 |
| **DVG-Face** | **100.0** | **99.2** | **97.3** |

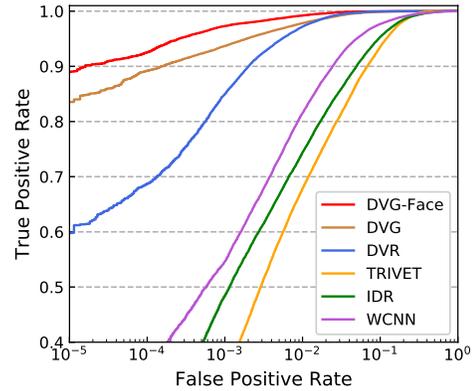

Fig. 1: ROC curves on Oulu-CASIA NIR-VIS.

TABLE 2: Experimental results under $\pm 90^o$ on MultiPIE.

| Method | Rank-1 |
|--------|--------|
| GridFace [3] | 75.4 |
| FF-GAN [4] | 61.2 |
| TPGAN [5] | 64.6 |
| CAPG-GAN [6] | 66.1 |
| PIM [7] | 86.5 |
| Rotate-and-Render [8] | 94.4 |
| DVG | 83.9 |
| **DVG-Face** | **97.6** |

# 1 EXPERIMENTS

## 1.1 Databases and Protocols

*Oulu-CASIA NIR-VIS Database* is a widely utilized NIR-VIS database that contains 80 identities with various poses and expressions. 50 identities are from the University of Oulu and the other 30 identities are from CASIA. Following the protocol of [1], only 40 identities are used in experiments, with 20 used for training and 20 used for testing. There are 48 NIR images and 48 VIS images for each identity. For testing, all of the VIS images from the 20 identities are adopted as the gallery set and the corresponding NIR images are used as the probe set. Evaluation indicators include the Rank-1 accuracy, VR@FAR=1%, and VR@FAR=0.1%.

*Multi-PIE Database* is often employed for multi-pose evaluations. There are 4 sessions with a total of 337 identities, and each identity has abundant illuminations, poses, and expressions. According to Setting 2 of [2], we only leverage the face images with the neutral expression under 20 illuminations. The first 200 identities are used for training and the remaining 137 identities are used for testing. For testing, the gallery set has one frontal image pair per identity, and the probe set consists of $\pm 90^o$ face images. The Rank-1 accuracy is reported.

*NJU-ID Database* is built to study ID-Camera face verification. This database has a total of 256 identities, and each identity contains one ID card image with 102×126 resolution and one camera image with 640×480 resolution. Since the NJU-ID database only has a limited number of images, we merely use it as the testing set, and collect a new ID-Camera database as the training set that has 1,190 image pairs. For testing, the camera images are used as the probe set, and the ID card images are used as the gallery set. The Rank-1 accuracy, VR@FAR=1%, and VR@FAR=0.1% are reported.

## 1.2 Comparisons with State-of-the-Art Methods

### 1.2.1 Results on the Oulu-CASIA NIR-VIS Database

Different from the CASIA NIR-VIS 2.0 database, the training set of the Oulu-CASIA NIR-VIS database only has 20 identities. Such a low-shot database poses huge challenges for the training of the HFR network. We evaluate our method against 5 traditional methods and 4 deep learning based methods. The compared traditional methods include MPL3, KCSR, KPS, KDSR, and H2(LBP3). The compared deep learning based methods include TRIVET, W-CNN, PACH, and DVR.

In Table 1, it is observed that the deep learning based methods significantly outperform the traditional methods in terms of the Rank-1 accuracy. At the same time, we also notice that compared with the Rank-1 accuracy, verification rates of all of the methods are not particularly high. This may be caused by the testing protocol, in which there are merely 20 identities in the testing set, and each identity contains up to 48 VIS images as the galleries and 48 NIR images as the probes. As a consequence, it is relatively

TABLE 3: Experimental results on NJU-ID.

| Method | Rank-1 | VR@FAR=1% | VR@FAR=0.1% |
|---|---|---|---|
| LightCNN | 91.3 | 90.5 | 80.4 |
| DVG | 96.8 | 96.7 | 83.0 |
| **DVG-Face** | **98.4** | **98.1** | **91.2** |

TABLE 4: Verification rates on BUAA-VisNir under different parameter values.

| (a) $\lambda_1$ in Eq. (14) | | (b) $\lambda_2$ in Eq. (14) | | (c) $\eta$ in Eq. (15) | | (d) $\alpha$ in Eq. (20) | |
|---|---|---|---|---|---|---|---|
| $\lambda_1$ | FAR=0.1% | $\lambda_2$ | FAR=0.1% | $\eta$ | FAR=0.1% | $\alpha$ | FAR=0.1% |
| 0 | 97.0 | 0 | 95.3 | 0 | 96.6 | 0 | 95.1 |
| 10 | 98.6 | 0.10 | 97.6 | 0.05 | 98.6 | 2e-4 | 98.3 |
| 25 | 98.9 | 0.25 | 98.9 | 0.10 | **99.1** | 5e-4 | 98.9 |
| 50 | **99.1** | 0.50 | **99.1** | 0.20 | 98.7 | 1e-3 | **99.1** |
| 75 | 98.8 | 0.75 | 98.7 | 0.30 | 97.6 | 2e-3 | 98.4 |
| 100 | 98.4 | 1.00 | 98.4 | 0.40 | 97.0 | 3e-3 | 97.2 |

TABLE 5: Rank-1 accuracies on MultiPIE under different parameter values.

| (a) $\lambda_1$ in Eq. (14) | | (b) $\lambda_2$ in Eq. (14) | | (c) $\eta$ in Eq. (15) | | (d) $\alpha$ in Eq. (20) | |
|---|---|---|---|---|---|---|---|
| $\lambda_1$ | Rank-1 | $\lambda_2$ | Rank-1 | $\eta$ | Rank-1 | $\alpha$ | Rank-1 |
| 0 | 84.3 | 0 | 83.5 | 0 | 83.3 | 0 | 84.6 |
| 10 | 95.8 | 0.10 | 96.7 | 0.05 | 96.8 | 2e-4 | 97.1 |
| 25 | 96.9 | 0.25 | **97.6** | 0.10 | **97.6** | 5e-4 | 97.3 |
| 50 | **97.6** | 0.50 | **97.6** | 0.20 | 96.5 | 1e-3 | **97.6** |
| 75 | 96.3 | 0.75 | 96.7 | 0.30 | 95.9 | 2e-3 | 96.7 |
| 100 | 96.1 | 1.00 | 96.3 | 0.40 | 94.8 | 3e-3 | 95.8 |

TABLE 6: Verification rates on NJU-ID under different parameter values.

| (a) $\lambda_1$ in Eq. (14) | | (b) $\lambda_2$ in Eq. (14) | | (c) $\eta$ in Eq. (15) | | (d) $\alpha$ in Eq. (20) | |
|---|---|---|---|---|---|---|---|
| $\lambda_1$ | FAR=1% | $\lambda_2$ | FAR=1% | $\eta$ | FAR=1% | $\alpha$ | FAR=1% |
| 0 | 96.5 | 0 | 95.3 | 0 | 96.9 | 0 | 91.3 |
| 10 | 97.7 | 0.10 | 96.7 | 0.05 | 97.8 | 2e-4 | 96.6 |
| 25 | **98.1** | 0.25 | 97.6 | 0.10 | **98.1** | 5e-4 | 97.4 |
| 50 | **98.1** | 0.50 | **98.1** | 0.20 | 97.1 | 1e-3 | **98.1** |
| 75 | 97.9 | 0.75 | 97.8 | 0.30 | 96.5 | 2e-3 | 97.3 |
| 100 | 97.3 | 1.00 | 97.6 | 0.40 | 95.7 | 3e-3 | 96.2 |

easy to retrieve the correct image according to the highest similarity score, leading to the generally high Rank-1 results. However, due to the large domain discrepancy and the small number of the training data, the discriminability of the learned features is limited. This results in the relatively low verification rates. In such a challenging case, our method still gets the highest performances. In particular, the best VR@FAR=0.1% is improved from 88.2% to 97.3%. The ROC curves in Fig. 1 further demonstrate the significant superiority of our method. Compared with the state-of-the-art DVR, TPR@FPR@10$^{-5}$ is improved by 29.2%. Moreover, regarding TPR@FPR@10$^{-5}$, we beat the preliminary version DVG by 5.5%. The dual generation results on Oulu-CASIA NIR-VIS are displayed in Fig. 2 (b).

### 1.2.2 Results on the MultiPIE Database

Since the recognition under extreme poses is still a challenging problem, we further evaluate our method on the representative MultiPIE database. The compared methods include 6 state-of-the-art methods: GridFace [3], FF-GAN [4], TPGAN [5], CAPG-GAN [6], PIM [7], and Rotate-and-Render [8]. All of these methods except for GridFace are based on conditional generation.

Table 2 suggests that our method outperforms all of the compared methods. The best Rank-1 accuracy under the most challenging $\pm 90^o$ is improved from 94.4% to 97.6%. Compared with the preliminary version DVG, we obtain an improvement of 13.7%. The generated paired images on the MultiPIE database are shown in Fig. 2 (e).

### 1.2.3 Results on the NJU-ID Database

With the popularization of face authentication systems, the ID-Camera recognition has become increasingly important. Therefore, we also evaluate our method on the representative ID-Camera database NJU-ID. The results of LightCNN, DVG, and DVG-Face are reported in Table 3. It is obvious that DVG-Face shows clear superiority over the preliminary version DVG in terms of the Rank-1 accuracy, VR@FAR=1%, and VR@FAR=0.1%. In particular, VR@FAR=0.1% is improved by 8.2%. We do not display the generated data because it may violate privacy.

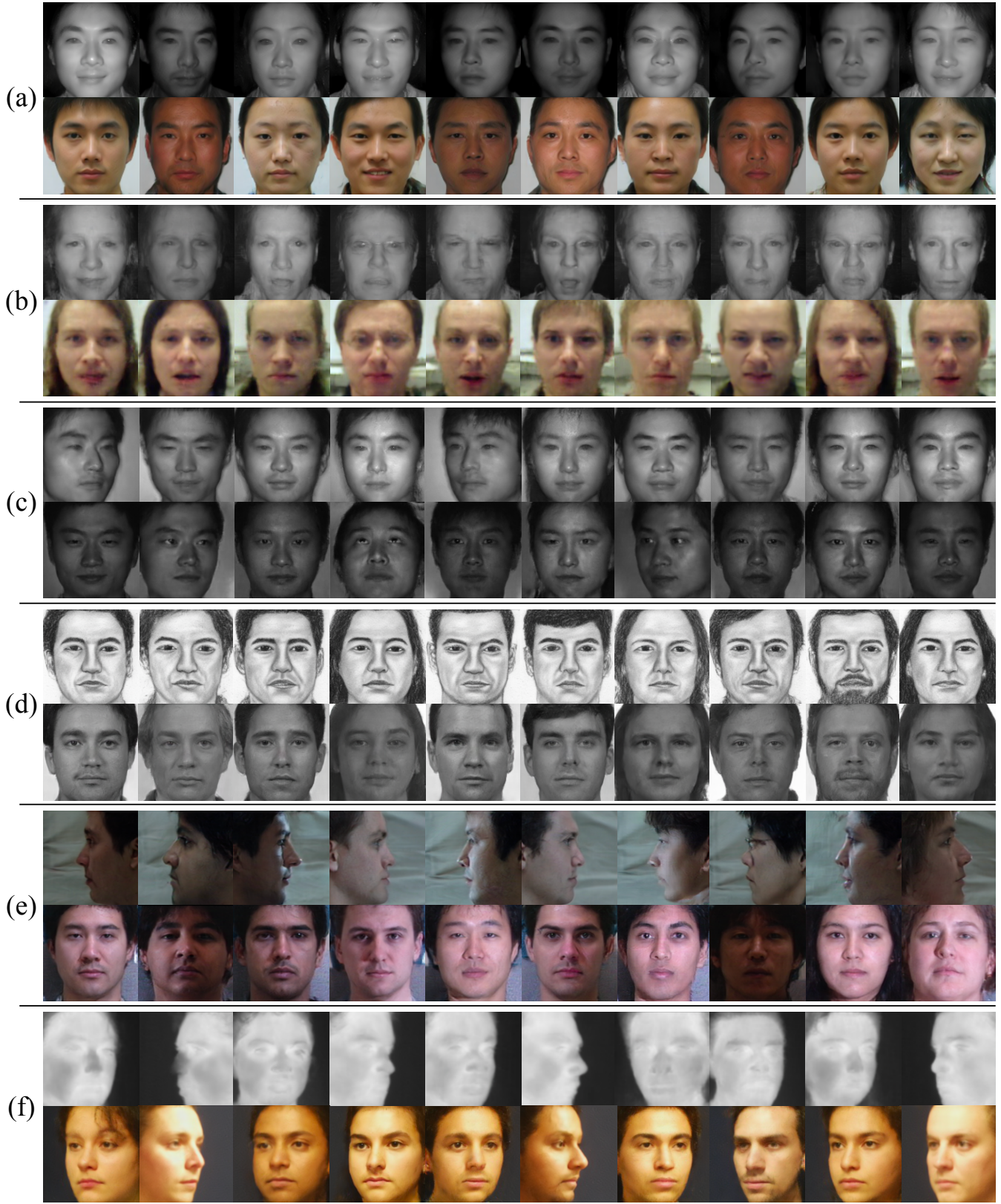

Fig. 2: Dual generation results on (a) CASIA NIR-VIS 2.0, (b) Oulu-CASIA NIR-VIS, (c) BUAA-VisNir, (d) CUFSF, (e) Multi-PIE, and (f) Tufts Face.



\end{document}